\documentclass[12pt,a4paper]{article}
\usepackage[hmargin={1.1in,1.1in},vmargin={1.1in,1.1in}]{geometry}
\usepackage[utf8]{inputenc}
\usepackage[T1]{fontenc}
\usepackage{amsmath}
\usepackage{amssymb}
\usepackage{graphicx}
\usepackage{hyperref}
\usepackage{natbib}
\usepackage{booktabs}

\usepackage{rotating}

\usepackage{pdflscape}

\usepackage{adjustbox}

\usepackage{booktabs}
\usepackage{multirow}
\usepackage{multicol}

\title{Inflation Attitudes of Large Language Models}

\author{Nikoleta Anesti,\thanks{\noindent Bank of England. Nikoleta.Anesti@bankofengland.co.uk} ~ Edward Hill,\thanks{\noindent Bank of England. Ed.Hill@bankofengland.co.uk} ~ Andreas Joseph\thanks{\noindent Corresponding author. Bank of England. Andreas.Joseph@bankofengland.co.uk. \newline \noindent We would like to thank conference participants at the ECB Macroeconomic Modelling Frontiers for Research and Policy in Central Banks, the 2024 Annual CEBRA, the 2024 ECONDAT Fall meeting, the 2025 Banque du France AI methods conference, the 2025 Annual RES meeting, and the 2025 BIS AI Summit. The views expressed in this paper are ours, and not necessarily those of the Bank of England or its committees. All errors are our own.}}
\date{\today.}
\begin{document}
\maketitle
\begin{abstract}
This paper investigates the ability of Large Language Models (LLMs), specifically GPT-3.5-turbo (GPT), to form inflation perceptions and expectations based on macroeconomic price signals. We compare the LLM's output to household survey data and official statistics, mimicking the information set and demographic characteristics of the Bank of England's Inflation Attitudes Survey (IAS). Our quasi-experimental design exploits the timing of GPT's training cut-off in September 2021 which means it has no knowledge of the subsequent UK inflation surge. We find that GPT tracks aggregate survey projections and official statistics at short horizons. At a disaggregated level, GPT replicates key empirical regularities of households' inflation perceptions, particularly for income, housing tenure, and social class. A novel Shapley value decomposition of LLM outputs suited for the synthetic survey setting provides well-defined insights into the drivers of model outputs linked to prompt content. We find that GPT demonstrates a heightened sensitivity to food inflation information similar to that of human respondents. However, we also find that it lacks a consistent model of consumer price inflation. More generally, our approach could be used to evaluate the behaviour of LLMs for use in the social sciences, to compare different models, or to assist in survey design.
\end{abstract}

\newpage

\section{Introduction}
Agent expectations have been a crucial component of many approaches in economic analysis, at least since  \cite{Lucas1975}. Economic agents who are forward looking, and the systems comprising them, behave substantially differently from those which merely react to current or past observations. However, the details of real-world expectations formation processes, like those of households and firms, are not yet well understood. They likely exhibit considerable heterogeneity \citep{DAcuntoWeber2024}, and common modelling approaches like rational or adaptive expectations each face challenges.\\
The release of ChatGPT in the end of 2022 led to a surge in the interest and the development of large language models (LLMs) and their applications. LLMs have shown impressive results across a variety of tasks, often reaching or exceeding human performance levels. Examples are the abilities to converse and reason, recall knowledge, answer logical questions, or write computer code \cite{christie2024judgement,galatzer2024benchmark,luo2024neuro,licorish2025comparing}. In short, LLMs can behave like, and master tasks comparable to, humans. Economics is no exception to this development. Early work uses LLMs as simulated economic agents who are given endowments, information and preferences so that their behaviour can be studied in various scenarios via simulations \cite{Horton2023}.\\ 
We contribute to this nascent literature on using LLMs for economic analysis by addressing the problem of understanding agents' economic perception and expectation formation, and by providing general approaches for analysing LLM outputs in the treatment setting. In particular, we investigate the ability of a version of OpenAI's GPT model\footnote{Slightly confusingly, the series of GPT models \citep{brown2020language} created by OpenAI, and used to power ChatGPT, are themselves examples of a generative pre-trained transformer (GPT), the generic term for the core technology underlying modern LLMs. We will use GPT or LLM interchangeably as representing this larger class of models. \textbf{Explicitly, we are not specifically discussing the use, advantages or disadvantages of OpenAI's models versus other flagship large language models, or endorsing, or not, their use. Rather the model in this work is used as a representative model  with which to test our methods and provide a subject for analysis.}} to assess consumer price inflation in the present and future when provided with different price signals. Our contributions can be summarised in four distinct points.\\
% We also present well-defined approaches to connect prompt content and model output, which can be of use beyond the current application.
%
One, we use a quasi-experimental design around a real-word scenario probing out-of-sample and out-of-distribution model behaviour. Following the synthetic survey setting \cite{Argyle2023simulate,arora2023probing}, we replicate two samples of the Bank of England's Inflation Attitudes Survey (IAS,  a quarterly survey which tracks inflation perceptions and expectations of UK households) around the peak of consumer price inflation in late 2022. While relatively brief, this shock to consumer prices was unseen alongside several dimensions. We investigate how GPT reacts to such an extreme event testing the limits of the model, but also how information leakage from this event may have affected subsequently released models.\\
Two, we provide novel machine learning interpretability tools for LLMs, which allow us to measure the effects of multiple information treatments and account for their interactions. We frame our work in the experimental treatment context. An LLM prompt consists of a synthetic persona based on the demographic characteristics of real-world survey respondents, economic conditioning information (treatment or a scenario), the IAS survey question, and instructions.\\
The discrete nature of the information treatment bits allows us to formulate a Shapley value decomposition \citep{StrumbeljKononenko2010} of the LLM's survey responses, so leveraging one of the most widely used and accepted machine learning explainability tools. The synthetic survey setting with information treatments is thus particularly well suited for this approach.\\
Three, we provide a multi-way evaluation of synthetic survey responses on the micro-, demographic meso-, and aggregate macro-level between simulated responses, actual responses, and official statistics. The IAS will serve as our ``human benchmark'' for model tuning and testing. The results are encouraging for the use of LLMs in our context. Aggregate response distributions can be matched quite well with the help of temperature tuning. On the meso-level GPT's outputs are often aligned with survey results along demographic groups, and are seen to be closer to official statistics than human responses. Interestingly, we find that GPT exhibits human-like biases including an oversensitivity to salient inflation components.\\
However, we also find reasons for caution when using LLMs. The micro-level correspondence between LLM and human responses is rather weak and partially unstable. This could be caused by the rudimentary economic conditioning environment and leaves plenty of scope for future research. Furthermore, the LLM we use exhibits inconsistencies which point to a lack of internal logic or the missing of a consistent world model of the concepts studied in this paper.\\
Four, we provide a discussion around the use of LLMs to inform decisions. Researchers and decision makers are ultimately interested in the behaviour of actual humans and so results from artificial intelligence (AI) experiments always require empirical validation. We provide approaches to this, but also highlight ethical considerations. LLMs can be biased, for instance with respect to demographic characteristics which we are investigating here \cite{bai2025bias}. In our setting, the situation can arise where LLM outputs match either the human benchmark or official statistics, but not both. When using LLM outputs in downstream tasks the original purpose of the analysis should be considered, and a strategy to handle the inevitable trade-offs devised.\\
Overall, we believe that our experimental design and approach consisting of validation, sensitivity analysis, and model explainability, all in the context of a three-way comparison between LLM, human, and official data, can be easily transferred to both other applications and different models, and as such contributes to an accepted framework for using LLMs in social science contexts. For instance, having studied a particular model and being aware of its strengths and weaknesses then opens the door for its use within the limits drawn out by the analysis, or may raise red flags for when a model cannot be used for a specific task. \\
Hurdles for and concerns about the use of LLM for research and decision making will likely need addressing for each situation separately, because, by their nature, the precise behaviour of a LLM cannot be known a priori in a given situation. Once overcome, one may (cautiously) use a model for the simulation of human subjects in social science, economics, or financial settings. Possible applications include the generation and cross-check of official statistics, including ``counterfactual statistics'', and the more efficient and effective design of surveys.\\
The remainder of this paper is structured as follows: Section\ \ref{sec:literature} summarises the literature on household inflation expectations, and the nascent use of LLMs in economics and surveys. Section\ \ref{sec:methods} describes the experimental setting and introduces the methodology. Section\ \ref{sec:results} presents the main results. We conclude with the discussion in Section\ \ref{sec:discussion}.

\section{Literature}
\label{sec:literature}

\subsection{Household Inflation Expectations and their determinants}

Inflation expectations play a special role in many areas in economics. For instance, in designing monetary policy because they determine households' savings and consumption decisions via the consumption Euler equation. Inflation expectations also drive agents' wage bargaining, durable investment including housing and mortgage choices, and portfolio choices \citep{Bernanke2007}. Because inflation expectations affect the decisions and actions of many economic actors, they also affect aggregate economic outcomes. Central Banks around the world actively try to manage inflation expectations, and understanding their key determinants and how they transmit into economic decisions is a topic of high policy relevance.\\
The key building block, across both theoretical and empirical strands, is the role of information frictions faced by households, with particular focus on financial literacy levels \citep{de2011measuring}, cognitive abilities \citep{d2019cognitive}, levels of attention \citep{sims2010rational,cavallo2017inflation}, sources of information \citep{lamla2015information}, subjective model of the economy \citep{macaulay2022heterogeneous}, transmission of policy communication \citep{coibion2022monetary,COIBION2020103297,d2020effective,ehrmann2022central,mcmahon2023getting} and personal inflation experiences.\\
Within that, most of the literature so far points to the significant role of everyday price signals observed by individuals \citep{MankiwReis2002, MackowiakWiederholt2009, CoibionGorodnichenko2015a}. Households also focus on the price changes of goods they purchase frequently, such as grocery items, rather than the price changes of a representative consumption bundle when forming their inflation expectations \citep{VanderKlaauw2012, deBruin2011, DAcunto2021}. Moreover, households tend to put a higher weight on positive than negative price changes when forming inflation expectations, which helps explain the persistent upward bias that has been documented in the literature \citep{Mankiw2003}.\\
Over the years, a growing literature has additionally focused on documenting and explaining empirical regularities of households' inflation expectations. The most salient feature is the substantial cross-sectional dispersion, which is systematically correlated with a set of demographic characteristics \cite{arioli2017eu,del2008s,jonung1981perceived}. This finding provides support to the argument that highlights the importance of individual-level drivers of inflation expectations, and suggests that traditional models of beliefs formation, which target the mean, median, or otherwise representative household expectations, fail to account for the most notable empirical regularities of the inflation expectations. For example, \cite{hobijn2009household} and \cite{kaplan2017inflation} study the variation in personal inflation rates experienced between households, with the latter documenting higher inflation rates among lower-income families. %Meanwhile, significant heterogeneity in inflation expectations across demographic groups is also well-documented in \cite{arioli2017eu,del2008s,jonung1981perceived}.
\\

\subsection{LLMs in economic analysis}

Like the personal computer, AI is special in the sense that it affects the structure of the economy but also provides a tool for analysis. The former is an active field of research (e.g. \cite{Acemoglu2022labour,hui2024labour,chen2024displacement}). For us, AI, in the form of LLMs, is a tool which can be used for economic research \cite{korinek2023ai,korinek2025ai,charness2025next}.\\
LLMs cannot only be used as tools to perform tasks as summarising literature, writing code, or helping with ideation, but also to simulate economic subjects themselves \cite{Argyle2023simulate,arora2023probing,manning2024automatedsocialsciencelanguage}. A range of work supports the idea that on individual, self-contained questions, value judgements and actions taken by LLMs align with human behaviours across psychological, philosophical, economic and political tests \citep{Aher2022, Brookins2023, FariaeCastro2023}. \cite{Horton2023} is an example of early work simulating economic agents, ``homo silicus''. Human behaviour often deviates from rational choice theory as investigated by behavioural economics. In light of this, it is of interest whether LLMs and the agents they represent behave rationally or rather ``human-like'' bounded-rationally. Early evidence suggests that it may depend on the context and the information provided \cite{ross2024llmeconomicusmappingbehavioral,henning2025llmagentsreplicatehuman}. At the same time, several studies show that LLMs know more than their human-readable outputs implies \cite{orgad2025llmsknowshowintrinsic,buckmann2025improving,buckmann2025}, suggesting that researchers may not yet have learned to fully utilise LLMs in different contexts. It has also been shown that LLMs memorise information selectively and partially from their training data \cite{lopezlira2025memorizationproblemtrustllms,crane2025}, making a clean out-of-sample evaluation essential for valid downstream inference \cite{ludwig2025llm}.\\[.2cm]
We will be using LLMs in an experimental setting simulating human survey subjects using conditioning information in a macroeconomic context \cite{Bybee2023,Hansen2025Simulating,Zarifhonarvar2025survey}. Our findings relate to the strand of the literature that examines the importance of personal inflation experiences via observed price signals, meaning that we will focus on the items like food and energy that households are exposed to more frequently, and will try to establish whether there is any association with simulated survey inflation expectations. The evidence so far has suggested that grocery prices bias inflation expectations systematically to the upside \cite{DAcunto2021}. The rationale being that grocery prices are more volatile than other prices and price changes typically revert more quickly. These features are one reason why many central banks focus on measures of core inflation, which exclude the price changes of groceries (and gas and energy), when examining inflationary pressures in the economy. However, by doing so, central banks risk overlooking inflation expectations picking up due to households' frequent exposure to higher than usual grocery and energy prices. The techniques presented below offer a laboratory to assess this for a given scenario.

\section{Methodology}
\label{sec:methods}

\subsection{Prompting Strategy}
We use OpenAI's GPT to answer the Bank's Inflation Attitudes Survey (IAS) in the context of the 2022 inflation surge, which peaked in October 2022. Specifically, we use OpenAI's \texttt{chat completions} application programming interface (API) to query its GPT models (specifically \texttt{gpt-3.5-turbo-0613}) programmatically creating 'synthetic IAS samples'. For price measures, we focus on consumer price index inflation including owner occupiers' housing costs (CPIH) and its subcomponents in the UK.\footnote{This is a more comprehensive measure than consumer price index inflation excluding owner occupiers housing costs (CPI). Both measures peaked at the same time in 2022.}\\
As has been proposed in previous work \citep{Jiang2022}, LLMs can be conditioned on representing a particular political or demographic group. We will include gender, age, income, housing tenure, social class, UK region in our analysis which are collected on an individual bases as part of the IAS (see Appendix \ref{app_demo_groups}). Additionally, we introduce economic conditions in the form of inflation of sub-components of the CPIH, in particular food,  restaurants \& cafes, energy, and everything else (other).\footnote{The first three correspond to the components 01, 11.1.1, 04.5 of the CPIH, respectively. Together these have about 21\% of index weight and the remainder is covered by the other component. In much of our analysis, we will treat food \& restaurants jointly.} More precisely, we provide the three-month average of year-on-year (yoy) inflation of each component preceding the survey month to approximate the information set actual survey respondents have with respect to consumer price inflation. The first three components are often more volatile and seen as more salient by households, who overweight them when forming inflation expectations \citep{repec:boe:boeewp:1125}. We will investigate whether GPT shows biases to any of the components we use for economic conditioning. Iterating through the survey sample at a given point in time will return UK representative synthetic sets of survey responses on inflation perceptions and expectations and allow us to investigate their drivers.\\
Perceptions here relate to the current rate of price inflation and expectations to expected year-on-year price changes in one, two, and five years.\footnote{Inflation perceptions can be seen as short-term expectations given the lag of six to seven weeks until price data are available after the end of the reference month.} We instruct the API through both the \texttt{system} and \texttt{user} prompts. The \texttt{system} prompt remains the same in all cases. An example prompt for inflation perceptions is the following:\\[.4cm] 
\texttt{System: You are pretending to be the person described given your best guess as to their personal, social and economic situation.}\\[.4cm] 
We then alter the \texttt{user} prompt as\\[.4cm]
\texttt{You are male, aged 16-24, live in the North of England or Northern Ireland, are upper-middle class and are not working with an income of >£45000. You got your A-levels but not a degree and live in a house you rent.}\\[.2cm] 
\texttt{In the last few months, food inflation has been 17\% (9.8\% in restaurants and cafes), energy price inflation was about 88\%. On average the rate of inflation on other goods was about \%}\footnote{Numbers bigger than ten in absolute terms are rounded to the next integer. One digit is given otherwise.}\\[.2cm] 
\texttt{You are going to be asked questions about your perception of current and future inflation.
Which of these options best describes how prices have changed over the last 12 months?}\footnote{Answer options are mapped to the midpoint of each interval or 0.5 percentage points beyond the reference value. For example, ``risen by more than 15\% '' is taken to be 15.5\%.}

\begin{enumerate}
\item \texttt{gone down by less than 1\%}
\item \texttt{gone down by 1-2\%}
\item \texttt{... [The other options in the IAS]}
\item \texttt{risen by 13-14\%}
\item \texttt{risen by 14-15\%}
\item \texttt{risen by more than 15\%}
\end{enumerate}
\texttt{Please choose one option, no explanation.}\\[.2cm]
LLMs have been shown to be sensitive to the order in which response choices are presented \citep{pezeshkpour2024large}. To address potential bias from a particular choice presentation, we scramble the response options for each respondent in each sample with its own random seed.

\subsection{Experimental Setting}

% >>> time series plot
\begin{figure}
\centering
\label{fig:time_series}
\includegraphics[width=1\textwidth]{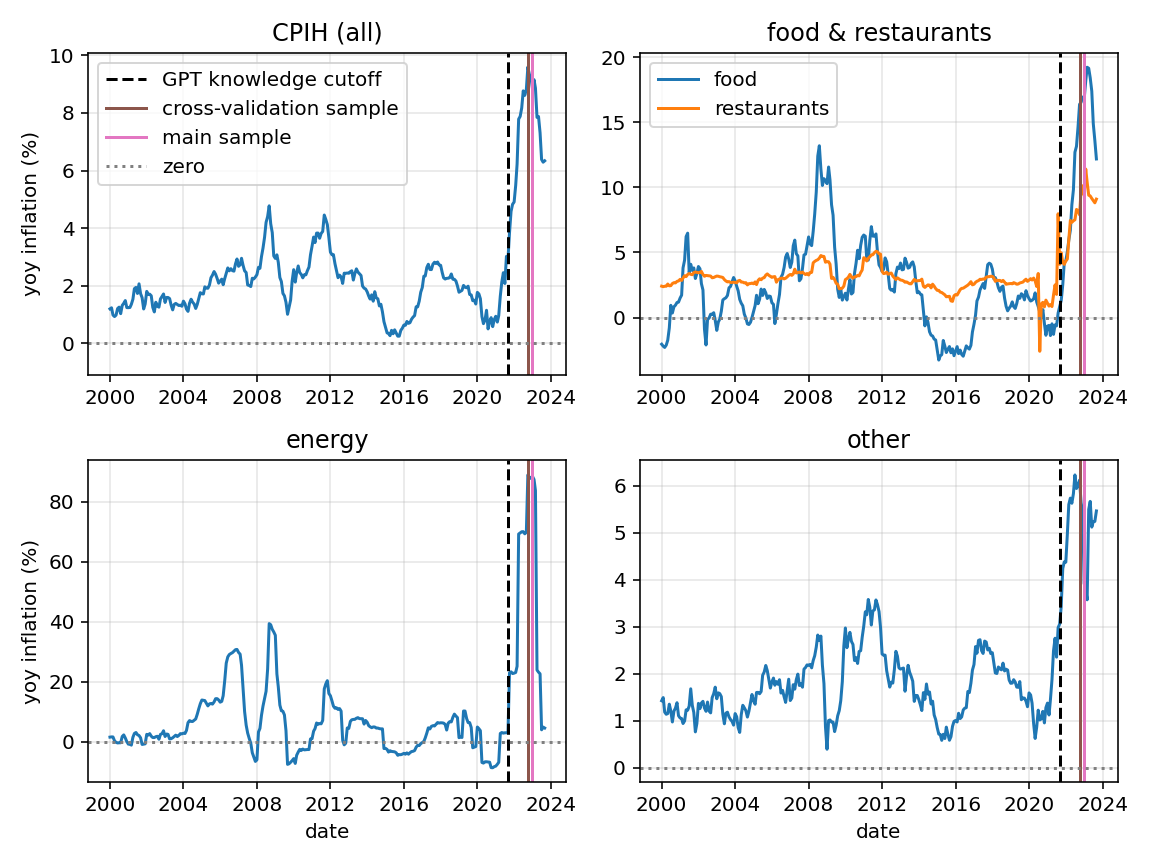}
\caption{Time series of CPIH year-on-year inflation and components. Vertical lines indicate GPT knowledge cut-off (dashed) and IAS survey samples used in November 2022 (cross-validation; 2022Q4) and February 2023 (main; 2023Q1). Sources: ONS.}
\end{figure}

In terms of survey timing, and the corresponding samples to concentrate on, we consider two samples at the peak of consumer price inflation in the end of 2022 and early 2023, namely 2022Q4 (November 2022) and 2023Q1 (February 2023). There are three main reasons for this. One, this brings us outside GPT's training period which ends in September 2021. As such it does not know about the following inflation surge or particular drivers contributing to it, like the Russian invasion of Ukraine and the subsequent spike in the energy prices among others\footnote{The Appendix contains a set of validation questions used to verify GPT's knowledge cut-off, and test whether there has been leakage into the model throughout the analysis. The answers to these questions have been stable over time with the latest test performed on 1.\ December\ 2025.}. This means that an information treatment (conditioning of GPT) related to the subsequent inflation surge can be interpreted as quasi-experimental. Two, this is around the time that aggregate inflation measures peaked but there still was uncertainty about their actual paths and how temporary this shock might have been. This means we are able to gauge the maximum impact of the inflation shock. Third, the size of the shock, as measured by economic conditioning information, was well outside the data ranges GPT would have observed in the past. This allows us to probe the model in a real, yet extreme, situation, and to evaluate its limits in the current context.\\
The experimental setting is depicted in Figure~\ref{fig:time_series}, which shows year-on-year inflation of the overall index, and the subcomponents we use in our analysis. The end of GPT's training period is given by the dashed vertical line where all inflation measures have still been well within historical ranges. The two solid vertical lines correspond to the two survey samples we use. These coincide with the peaks of the different series. The details of the economic conditions used in either sample are given in Table~\ref{tab:temp_cv}. These are broadly similar to each other, so we can feel confident carrying over insights gained in one sample to the other.\\ 
% To-DO: Third, the Q1 rounds of the IAS contain additional information about people's beliefs about the underlying inflation drivers. As such, they allow us to look into more details in the comparison between actual people and LLM representations.
%
In line with common practice in statistical learning, we use the first sample (2022Q4) for cross-validation (CV) and the second sample (2023Q1) as our main test sample. In particular, we will calibrate GPT's temperature parameters in the range $T\in\{0,0.25,0.5,0.75,1,1.25,1.5\}$.\footnote{Theoretically OpenAI's API allows to increase the temperature up to $T=2$. However, for $T>1.5$ we observe that GPT often is either not able to follow the instructions returning random strings unrelated to the prompt, or throws an exception after a considerable delay. We consider this a model breakdown making it basically unusable.} This affects how deterministic (smaller $T$) or random (higher $T$) GPT's answers are by affecting the width of its softmax output probability distribution. This gives us some control over the moments of GPT's response distribution.

\begin{table}[htbp]
\centering
\begin{tabular}{lcrrrr|c}
\toprule
sample & ref. times & food & restaurants & energy & other & CPIH\\
\midrule
cross-validation & Nov 2022 & 15.0 & 8.1 & 76.0 & 6.0 & 9.3 \\
main & Feb 2023 & 17.0 & 9.8 & 88.0 & 5.0 & 9.2 \\
average & Jan'89 - Sep'21 & 2.4 & 3.8 & 4.2 & 1.7 & 2.6 \\
\midrule
shares & Feb-Dec\ 2022 & 0.096 & 0.074 & 0.041 & 0.789 & 1.000 \\
\bottomrule
\end{tabular}
\caption{Economic conditioning information used in the different experiments. For the main and cross-validation samples, average sub-component year-on-year inflation the three month before the reference period is given; CPIH values are actually realised values at the reference times. Average values relate to those overlapping with the GPT training period pre-Oct 2021. Shares are the CPIH basket shares February-December 2022. Sources: ONS.}\label{tab:ec}
\end{table}

\subsection{LLM treatment effects}

One of the biggest advantages of the use of LLMs in the current context is that the analysis of treatment effects does not need to rely on the potential outcomes framework \cite{rubin2005outcomes}, which stipulates the impossibility of observing the treated and untreated at the same time. By virtue of simulating subjects, survey respondents in our case, we are always able to observe the same subject under any treatment state.\\
Let $t_i\in\{0,1\}$ refer to a subject $i$ receiving a treatment $t$ or not. Here, $t=0$ may be a reference treatment, like a placebo. In our case, $t=1$ corresponds to including the economic conditions for the two survey samples listed in Table~\ref{tab:ec} in the user prompt. No treatment, $t=0$, is either the omission of economic conditions in the prompt, or the inclusion of some reference values, e.g.\ the pre-training cut-off historic averages (see Table~\ref{tab:ec}). With $x_i$ being the vector of demographic characteristics of subject $i$, their {\it individual treatment effect} simulated by GPT can be written as
\begin{equation}
\label{eq:treatment}
\tau_i\; =\; g(t=1;x_i) - g(t=0;x_i)\,,
\end{equation}
where $g(\cdot)$ is the GPT output with the assumption that we can perform a meaningful difference operation. This will be trivial in our case as we map all survey responses to numbers. With a sample size $N$, the {\it average treatment effect} simulated by GPT is
\begin{equation}
\label{eq:ave_treatment}
\bar{\tau}_i\; =\; \frac{1}{N}\sum_{i=1}^{N}\, [g(t=1;x_i) - g(t=0;x_i)]\,.
\end{equation}
There are two major concerns regarding the validity of \eqref{eq:ave_treatment}. First, the sample over which it has been calculated. Second, potential bias coming from the use of LLMs instead of human subjects. The first is common to the treatment and survey literature and is addressed by generating a nationally representative sample stemming from the underlying IAS. Addressing the second is one of the contributions of the paper, where we will perform a three way comparison between GPT, the IAS, and official out-turns. 

\subsection{LLM explainability}

Machine learning models, including LLMs, are often subject to the black box critique: there are no clear input-output relations which can be used to explain a model's predictions based on its inputs. Such relations are simple to obtain in a linear regression model where a variable's coefficient is the measure of the input-output relationship. However, since machine learning models do not specify an explicit functional form, there is no corresponding concept of a coefficient, making model explanation, interpretation and investigation challenging. Additionally, the black box critique is particularly severe for LLMs because of the high-dimensional and unstructured nature of their inputs and outputs (such as text) and the fact that a user of a commercial LLM will not have direct access to a fitted model's (very many) parameters.\\

Shapley values are a well-established tool for explaining machine learning model predictions \citep{StrumbeljKononenko2010}. Shapley values are a concept borrowed from game theory, where they describe the contributions of players to a cooperative game's group payoff. In the modelling setting, they can be used to decompose model predictions based on the contributions from each input variable. This information can then be used to identify model drivers and potentially complex non-linear relationships learned by a model. The Shapley value for a feature $k$ and observation or subject $i$ for a model $g(\cdot)$ can be written as

\begin{equation}
\label{eq:shap}
    \tau_{i,k}^S\big(x_i;g\big) \,=\, \sum_{x'\,\subseteq\,\mathcal{C}(x)} \frac{|x'|!(K-|x'|-1)!}{K!}\,\big[g(x_i|x'_i\cup x_{ik}) - g(x_i|x'_i)\big]\,,
\end{equation}

where the variable set $x'$ runs over all sets $\mathcal{C}(x)\setminus k$, which is the set of all possible variable combinations of $K-1$ variables when excluding $k$. The combinatorial weighting factor $|x'|!(K-|x'|-1)!/K!$ sums to one over $\mathcal{C}(x)\setminus k$. Eq.\ \ref{eq:shap} can be interpreted as the marginal contribution of variable $k$ to all possible coalitions excluding it taking all possibilities of complementing or substituting any other individual or group of variables into account.\\

We propose a general framework to address the black box critique based on explainable machine learning approaches \citep{BuckmanJoseph2022}. In particular, we will adapt the Shapley value framework to the survey and treatment setting \citep{Joseph2019}. The application of Shapley values to LLMs in the general case is difficult, because of the difficulty parsing inputs into discrete variables. While general text inputs are encoded into lists of discrete variables using byte-pair encoding (BPE, \cite{brown2020language}) and we could perhaps extract active tokens from that through considering semantics or syntax, any approach will be complex in itself and quickly run into the curse of dimensionality given the computational complexity of $K!$ in \eqref{eq:shap}. 

However, this situation is considerably simplified in the survey and treatment setting. The conditioning information (inputs) can be readily separated, ex ante, into discrete parts, e.g.\ demographic categories, and the response is a single number (inflation perceptions or expectations). This means that LLM predictions can be decomposed similarly to the conventional case of supervised learning with a single target to model.

An interesting question is what the relation between the Shapley value \eqref{eq:shap} of a treatment $t$ and its treatment effect is. Based on \eqref{eq:treatment}, these are the same for a single treatment. However, we have multiple treatments $t=(t_1,\dots,t_d)$ in our case corresponding to information on the several CPIH price components we include in the LLM's prompt, i.e.\ $t=(t_f,t_r,t_e,t_o)$ for the food, restaurants \& cafes, energy, and other components, respectively. A na\"ive way for evaluating a single treatment, say, the effect of high energy prices on inflation perceptions would be to set the remaining information treatment values to some neutral values, like long-run averages or null, or excluding it altogether, and then subtract that model prediction from the full treatment case.\\
The choice of the untreated or control reference depends on the question being answered. For example, if one wants to know how high energy prices affect inflation perception {\it all other things being normal}, one can take long-run averages $\bar{t}=(\bar{t}_f,\bar{t}_r,\cdot,\bar{t}_o)$ for the other treatment values and write down the {\it na\"ive treatment effect}

\begin{equation}
\label{eq:energy_naive}
\tau^n_{i,e}\; =\; g\big(t=(\bar{t}_f,\bar{t}_r,t_e,\bar{t}_o);x_i\big) - g\big(t=(\bar{t}_f,\bar{t}_r,\bar{t}_e,\bar{t}_o);x_i\big)\,,
\end{equation}
where the only difference between the two terms on the right-hand side is in the value of the energy information treatment. The above expression is na\"ive in the sense that for a particular multi-treatment scenario, we expect the joint set of inputs to matter, i.e.\ there are potentially important interactions between the individual treatment components. Exactly this situation is taken into account in the calculation of Shapley values in in \eqref{eq:shap} by the consideration of all possible subsets of variables not including the variable of interest, $t_e$ in the current example. Following \eqref{eq:shap}, the Shapley value for energy is calculated as 

\begin{align}
\label{eq:energy_shap}
\tau^S_{i,e}\; &=\; \frac{1}{3}\bigg[g\big(t=(\bar{t}_f,\bar{t}_r,t_e,\bar{t}_o);x_i\big) - g\big(t=(\bar{t}_f,\bar{t}_r,\bar{t}_e,\bar{t}_o);x_i\big)\bigg]\\
 &+\; \frac{1}{6}\bigg[g\big(t=(t_f,t_r,t_e,\bar{t}_o);x_i\big) - g\big(t=(t_f,t_r,\bar{t}_e,\bar{t}_o);x_i\big)\bigg]\\
 &+\; \frac{1}{6}\bigg[g\big(t=(\bar{t}_f,\bar{t}_r,t_e,t_o);x_i\big) - g\big(t=(\bar{t}_f,\bar{t}_r,\bar{t}_e,t_o);x_i\big)\bigg]\\
 &+\; \frac{1}{3}\bigg[g\big(t=(t_f,t_r,t_e,t_o);x_i\big) - g\big(t=(t_f,t_r,\bar{t}_e,t_o);x_i\big)\bigg]\,,
\end{align}
where we treated the food and restaurant \& cafes components as a single variable which is either active (scenario value) or passive (average value). This can be done because of the linearity of \eqref{eq:shap} allowing us to considerably reduce the computational burden by bunching variables.\footnote{The details regarding a low-dimensional subset of variables is often of interest in these high-dimensional settings, where one can either create lower-dimensional `factors' leveraging domain knowledge or an algorithmic approach, or take a small number of variables of interest and treat all others jointly as `other' as we do. Both approaches allow us to considerably reduce the computational complexity of \eqref{eq:shap} while still being exact. Further approximations can be made by sampling coalitions from $\mathcal{C}(x)$.} We will follow the bunching approach treating ``food'' and ``restaurant \& cafes'' as a single ``food \& restaurants'' variable. This also is an example of how Shapley values can be used to consistently represent complex quantities by the grouping variables. \\
We will compare results from the na\"ive and Shapley treatment evaluations and see that they can differ indicating important interactions between treatment subcomponents.

\section{Results}
\label{sec:results}

\subsection{Temperature calibration}
\label{sec:temperature}

\begin{figure}
\centering
\includegraphics[width=.7\textwidth]{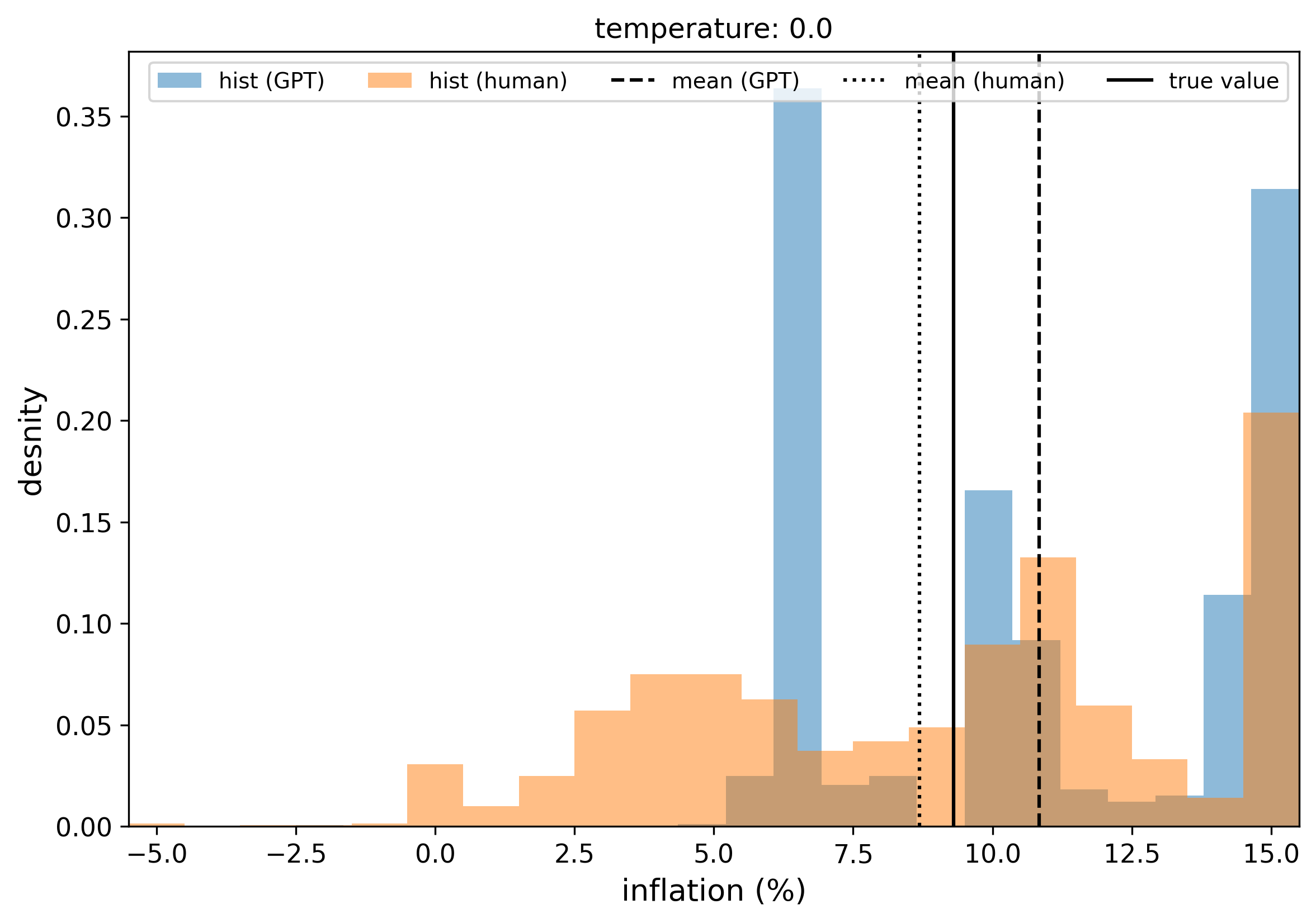}\\
\includegraphics[width=.7\textwidth]{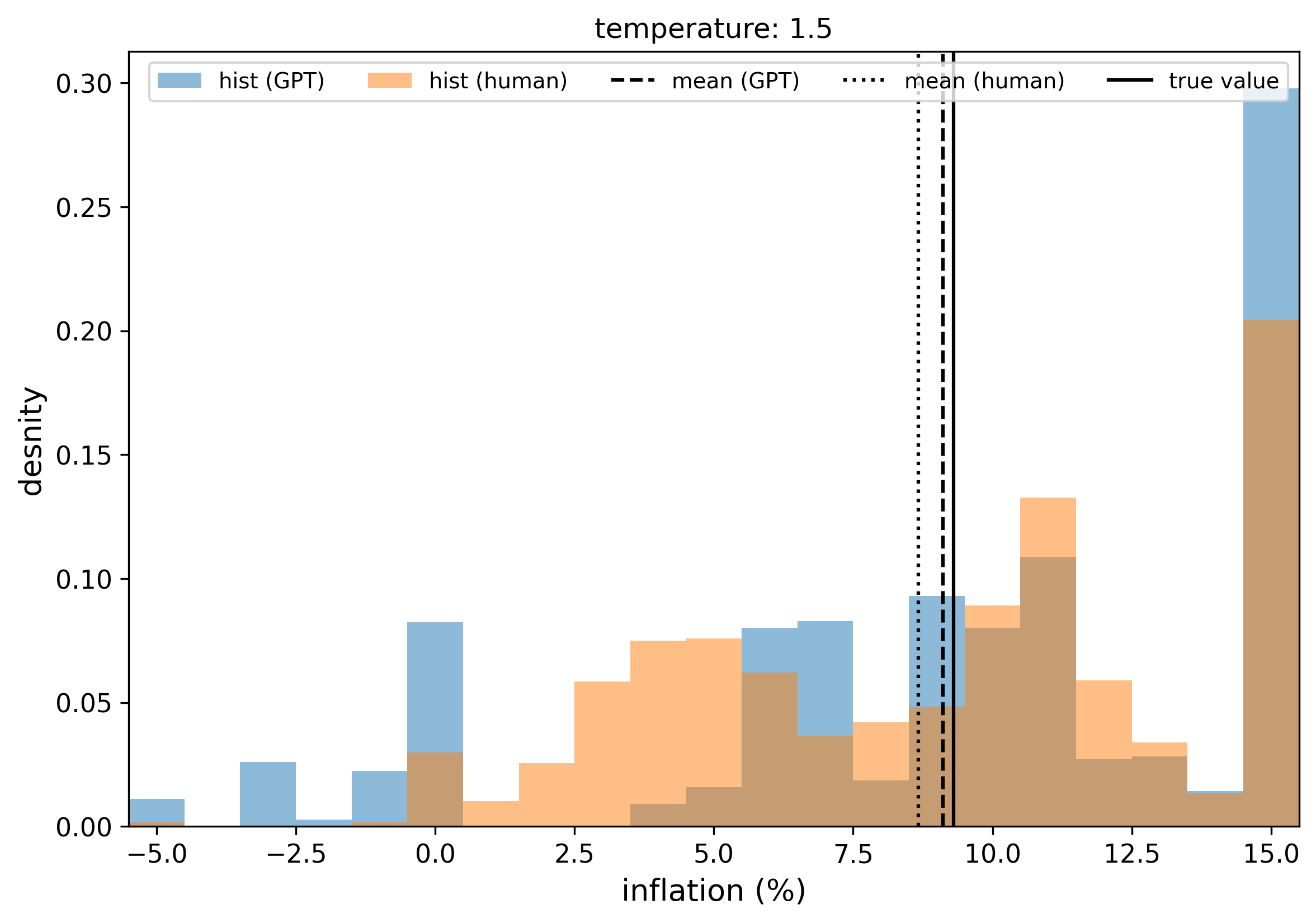}
\caption{Histograms of human (sand) and GPT (blue) responses for conditioned inflation perceptions in the cross-validation sample (November 2022) for $T=0$ (upper panel) and $T=1.5$ (lower panel). Sources: IAS, ONS,  authors' calculations.}
\label{fig:hist_CV}
\end{figure}

We investigate how the temperature parameter ($T$) affects GPT's response distribution using values in the range $T\in\{0,0.25,0.5,0.75,1,1.25,1.5\}$ for inflation perceptions in the IAS cross-validation sample (2022Q4). We track the mean and standard deviation of the resulting GPT response distributions and compare them to those of human responses. We summarise this comparison in the equally weighted loss function

\begin{equation}
\label{eq:loss}
L(y_{GPT},y_{IAS},w;l)\,=\,\frac{1}{2}\big|MN(y_{GPT};w)-MN(y_{IAS};w)\big|^{l}+\frac{1}{2}\big|SD(y_{GPT};w)-SD(y_{IAS};w)\big|^{l}\,,
\end{equation}
where $MN(\cdot,w)$ and $SD(\cdot,w)$ are weighted mean and standard deviation of the input vector with survey weights $w$, and $l\in\{1,2\}$ corresponding to a linear or a quadratic loss. This allows us to (try to) match aggregate survey responses for validation. We also consider the relation between GPT and IAS on the individual or micro-level by tracking the Pearson correlation coefficient between the two for different temperature values.\\
The results for this exercise are summarised in Table~\ref{tab:temp_cv}, with temperature values increasing from the top to the bottom. We make several observations. First, both loss measures decrease monotonically with increasing temperature, meaning that a higher temperatures leads to a better match of GPT to human responses. Second, GPT tends to predict higher inflation values than humans (the difference between means is always positive), while the width of GPT's response distribution surpasses that of human response for $T=1.25$ and above. This can be seen in Figure~\ref{fig:hist_CV} which shows histograms for GPT and IAS inflation perceptions for $T=0$ (upper part) and $T=1.5$ (lower part) for the cross-validation sample. Visually, GPT responses match IAS responses well for $T=1.5$. Additionally, mean GPT responses are also  close to the true value of aggregate consumer price inflation in November\ 2022.

% >>> CV table
\begin{table}[htbp]
\centering
\begin{tabular}{c|crcccccccc}
\toprule
 T & $n_{miss}$ & MN & diff$_{MN}$ & SD & diff$_{SD}$ & $L1$-loss & $L2$-loss & pcc & pval \\
\midrule
 0.00 & 0 & 10.83 & 2.15 & 3.80 & -0.97 & 1.56 & 2.77 & 0.05 & 0.02 \\
 0.25 & 0 & 10.80 & 2.11 & 3.82 & -0.94 & 1.53 & 2.67 & 0.05 & 0.03 \\
 0.50 & 0 & 10.73 & 2.05 & 3.91 & -0.85 & 1.45 & 2.46 & 0.04 & 0.10 \\
 0.75 & 0 & 10.62 & 1.93 & 4.18 & -0.59 & 1.26 & 2.04 & 0.04 & 0.06 \\
 1.00 & 0 & 10.25 & 1.56 & 4.54 & -0.22 & 0.89 & 1.25 & 0.03 & 0.15 \\
 1.25 & 1 & 9.93 & 1.25 & 4.87 & \phantom{-}0.10 & 0.67 & 0.78 & 0.05 & 0.02 \\
 1.50 & 49 & 9.11 & 0.44 & 5.58 & \phantom{-}0.81 & 0.63 & 0.42 & 0.05 & 0.04 \\
\bottomrule
\end{tabular}
\caption{Cross-validation statistics for GPT inflation perceptions for different temperatures ($T$): GPT missing values (NA responses), survey weighted mean (MN), difference to IAS mean, weighted sample standard deviation (SD), difference to IAS SD, $L1$-loss of GPT difference to IAS mean and SD (equally weighted), same $L2$-loss, Pearson correlation coefficient between GPT and IAS responses, and corresponding $p$-value. Sources: IAS, authors' calculations.}
\label{tab:temp_cv}
\end{table}

A third observation is that GPT-IAS micro-level correlations are stable but arguably quite low across the temperature range. This means that despite good matches on the aggregate level, GPT is not necessarily a good model for individual subject responses.\footnote{The response order randomisation on the individual level will be an additional reason for this, which we will not further investigate here.} This does, however, not mean that a lack of micro-level agreement means that higher-level results will not be accurate or useful. Collective or aggregate decision making has been shown to potentially be more accurate compared to individual estimates \citep{Galton1907VP,Krause2011SI}. We will, however, see that the results for the micro-level comparison may not be robust for high temperatures, and therefore continue our analysis by considering the cases of $T=1.5$ and $T=1$ side by side.

% >>> CV histogramms for T=0 and T=1.5

\begin{figure}
\centering
\includegraphics[width=.7\textwidth]{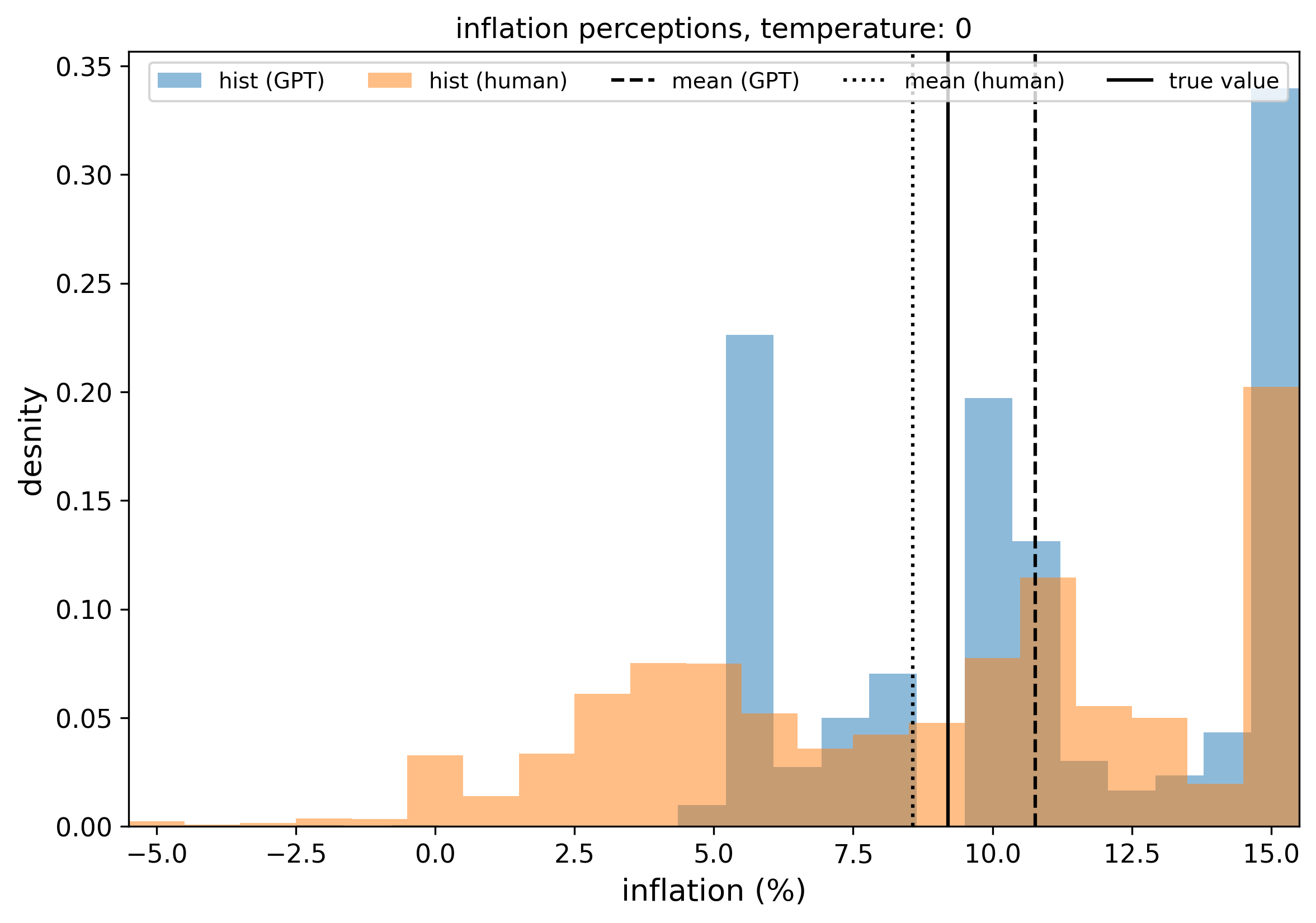}\\
\includegraphics[width=.7\textwidth]{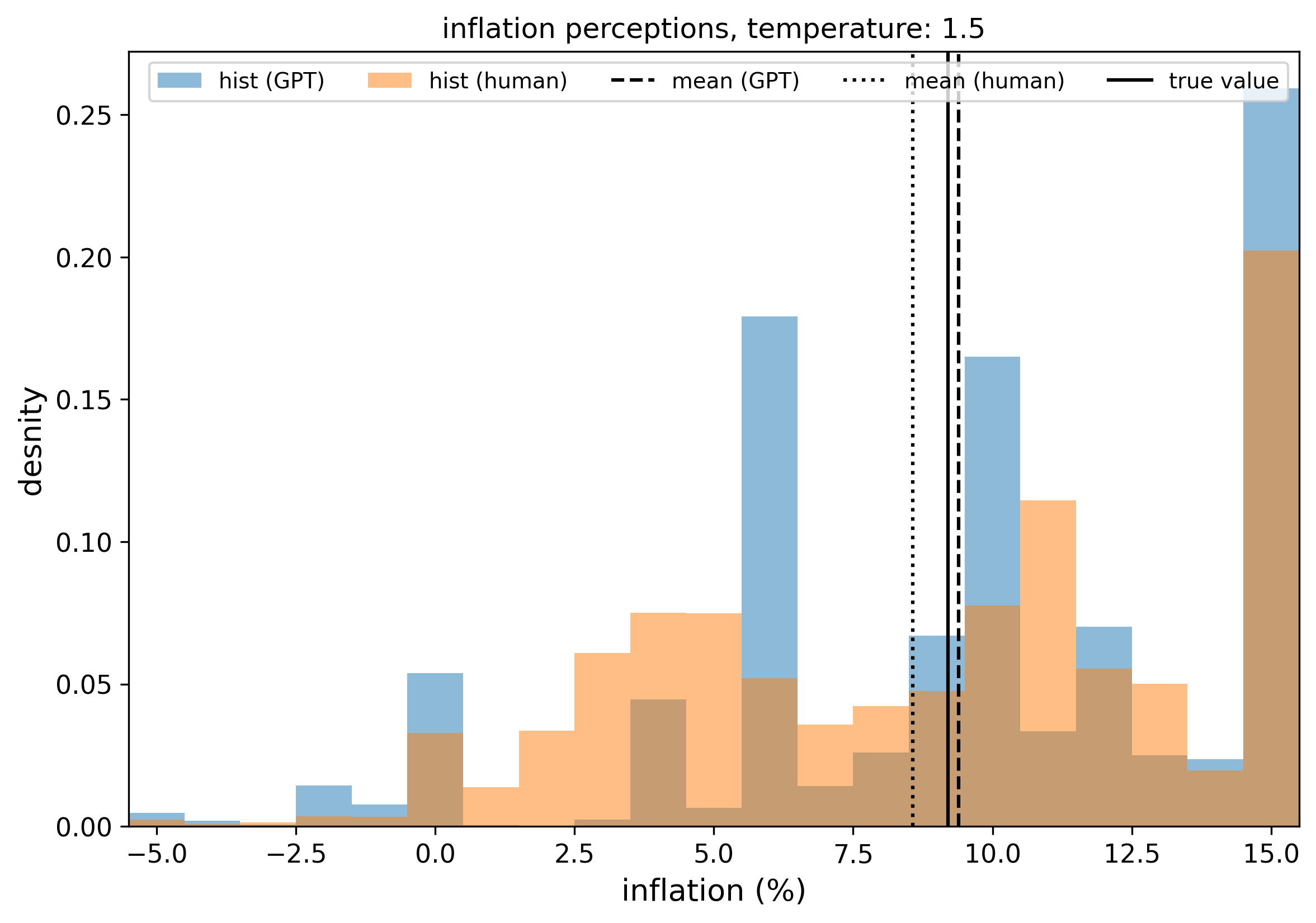}
\caption{Histograms of human (sand) and GPT (blue) responses for conditioned inflation perceptions in the main sample (February 2023) for $T=0$ (upper panel) and $T=1.5$ (lower panel). Sources: IAS, ONS, authors' calculations.}
\label{fig:hist_test}
\end{figure}

The corresponding test histograms for the main sample (2023Q1) are given in Figure~\ref{fig:hist_test}. We see again that the $T=1.5$ GPT response distribution overlaps well with IAS responses. Additionally, the mean value is again close to the actual realised official value (solid line). We will investigate these results in more detail in the next section alongside the response profile across time horizons. 

% >>> test histograom for T=1.5 and T=0

\subsection{Time profile of inflation expectations}
\label{sec:time}

For both economic theory and policy making, inflation expectations are paramount. A crucial question is whether these are `anchored' at about the central bank's inflation target on longer horizons. We investigate GPT inflation expectations time profile for the IAS 2023Q1 sample with the economic conditioning given in Table~\ref{tab:ec}. 

% >>> time profile plot
\begin{figure}
\centering
\includegraphics[width=.7\textwidth]{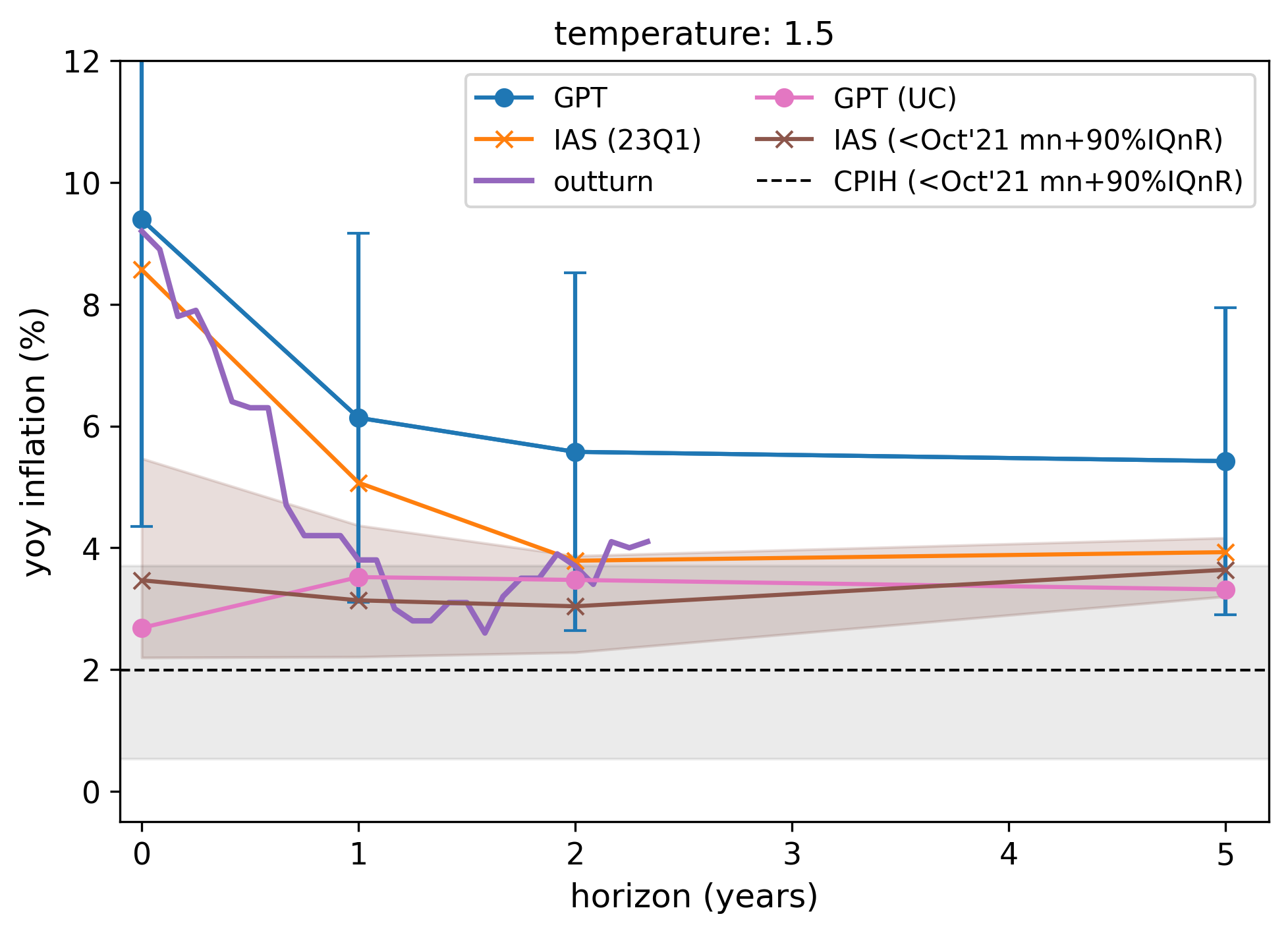}
\caption{Time profile of aggregate inflation expectations for IAS and GPT responses ($T=1.5$) for the main sample (2023Q1 survey): GPT conditioned (blue), GPT unconditioned (pink), IAS (sand), IAS pre-Oct-2021 mean \& swath (brown), out-turn from February 2023 onward (purple), and CPIH pre-Oct-2021 mean \& swath. Sources: IAS, ONS, authors' calculations.}
\label{fig:time_profile}
\end{figure}

The time profile of aggregate GPT expectations for horizons of up to five years is shown in Figure~\ref{fig:time_profile} for $T=1.5$.\footnote{The profile for $T=0$ is qualitatively very similar and is given in the Appendix.} The close match of inflation perceptions (horizon zero) between GPT, IAS, and ONS out-turn matches the lower part of Figure~\ref{fig:hist_test}. Both GPT (blue) and IAS (orange) expectations decrease with the horizon. However, there are a major discrepancies between future horizons from two to five years out. Though somewhat elevated, IAS mean expectations are in line with the historical distribution (IAS swath) and at the upper bounds of historically observed year-on-year inflation (CPIH swath).\footnote{Historical back data always end in Sep-21 coinciding with GPT's knowledge cut-off if not stated otherwise.} In contrast, GPT expectations stay roughly constant and elevated beyond the one year horizon. This suggests caution when using GPT infer inflation expectations. 

% >>> horizon stats for T=1.5 and T=0
\begin{sidewaystable}
\centering
\begin{tabular}{c|rrcrrccrr|ccrc}
\toprule
horizon & n$_{miss}$ & $MN$ & diff$_{MN}$ & SD & diff$_{SD}$ & L1-loss & L2-loss & pcc & pval & $MN_{uc}$ & effect$_{MN}$ & SD$_{uc}$ & effect$_{SD}$ \\
\midrule
\multicolumn{14}{c}{GPT ($T=1.5$)}\\
\midrule
0 & 13 & 9.39 & 0.82 & 5.04 & 0.10 & 0.46 & 0.34 & -0.01 & 0.49 & 2.68 & 6.71 & 3.23 & 1.80 \\
1 & 44 & 6.14 & 1.07 & 3.03 & -1.81 & 1.44 & 2.21 & 0.04 & 0.01 & 3.52 & 2.62 & 1.67 & 1.36 \\
2 & 84 & 5.58 & 1.79 & 2.94 & -1.62 & 1.71 & 2.92 & 0.02 & 0.18 & 3.47 & 2.10 & 1.69 & 1.25 \\
5 & 102 & 5.42 & 1.50 & 2.52 & -2.24 & 1.87 & 3.62 & 0.04 & 0.01 & 3.32 & 2.11 & 1.51 & 1.01 \\
\midrule
\multicolumn{14}{c}{GPT ($T=0$)}\\
\midrule
0 & 0 & 10.76 & 2.19 & 3.79 & -1.15 & 1.67 & 3.07 & 0.03 & 0.06 & 3.18 & 7.58 & 1.75 & 2.04 \\
1 & 0 & 6.26 & 1.19 & 2.09 & -2.75 & 1.97 & 4.48 & 0.05 & 0.00 & 3.23 & 3.03 & 1.13 & 0.96 \\
2 & 0 & 5.65 & 1.85 & 1.81 & -2.76 & 2.31 & 5.52 & 0.07 & 0.00 & 3.17 & 2.48 & 1.05 & 0.77 \\
5 & 0 & 5.36 & 1.42 & 1.76 & -3.01 & 2.21 & 5.53 & 0.07 & 0.00 & 3.10 & 2.26 & 0.97 & 0.78 \\
\bottomrule
\end{tabular}
\caption{Test statistics for GPT responses for different expectation horizons at $T=1.5$ (upper panel) and $T=0$ (lower panel): horizon in years (zero is current inflation perceptions), GPT missing values (NA responses), survey weighted mean ($MN$), difference to IAS mean, weighted sample standard deviation (SD), difference to IAS SD, $L1$-loss of GPT difference to IAS mean and SD (equally weighted), same $L2$-loss, Pearson correlation coefficient between GPT and IAS responses, corresponding $p$-value, weighted mean of unconditioned GPT responses, difference in mean GPT responses, weighted SD of unconditioned GPT responses, and difference in SD of GPT responses. Sources: IAS, authors' calculations.}
\label{tab:time_profile}
\end{sidewaystable}

Comprehensive summary statistics comparing GPT and IAS at different horizons and for a high ($T=1.5$) and low ($T=0$) temperature are given in Table~\ref{tab:time_profile}. The right part of the table investigates the effect of including economic conditioning information in the prompt compared to only including demographics. The effect columns show the average treatment \eqref{eq:ave_treatment} for the mean and standard deviation of the GPT response distributions. The inclusion of economic effects has shifted the mean considerably upwards. This was expected given that all treatment components have been considerably above their historic averages, such that this is a sense check.\\
Looking at the last column of Table~\ref{tab:time_profile}, we see that the width of the GPT response distribution has also increased as a consequence of the information treatment. This suggests that there are interactions between the demographics of the different personas given in the prompts and the economic conditioning. We will investigate this in the next section. However, we will first analyse the effect of temperature on GPT's responses in more detail. 

\subsubsection{High versus low temperature}

We see in Table~\ref{tab:time_profile} that there are considerable differences between the results for high and low GPT temperatures. In line with the cross-validation results, the losses \eqref{eq:loss} are lower for $T=1.5$. in almost all case, the higher temperature setting better matches both the mean and standard deviation of the IAS response distribution.\\
However, setting a high temperature also has considerable drawbacks. First, there can be a considerable number of invalid responses, in our case particularly for longer horizons. It is not clear what is driving this result. A possible explanation may be that the questions for longer horizons are logically more challenging as they refer to year-on-year changes after a certain time has passed. This may complicate following the instructions.\\
Second, the micro-level relation between GPT and IAS responses as measured by Pearson correlations is weak, volatile across horizons, and even partly breaks down. In contrast, the GPT-IAS relations are considerably stronger and actually increasing with the horizon for the zero temperature setting.\\
Lastly, the cross-horizon correlations are mostly weak for the high temperature case with patterns very different to human responses. The corresponding cross-correlations are listed in Table~\ref{tab:time_corr}. GPT at zero temperature shows patterns much more similar to those of human responses, especially between expectations at longer horizons.\\
Because of these observations, we will focus on the $T=0$ case below. This has the additional advantage that the interpretation of results does not carry uncertainty from the temperature setting which still can be investigated separately.

% >>> inter-horizon correlations (T=1.5, T=0, humans)
\begin{table}[htbp]
\centering
\begin{tabular}{l|cccc}
\toprule

\multicolumn{5}{c}{GPT (T=1.5)}\\
\midrule
$h$ & 0 & 1 & 2 & 5 \\
\hline
0 & 1.00 & 0.08 & 0.08 & 0.13 \\
1 & - & 1.00 & 0.82 & 0.20 \\
2 & - & - & 1.00 & 0.16 \\
5 & - & - & - & 1.00 \\
\midrule
\multicolumn{5}{c}{GPT (T=0)}\\
\midrule
$h$ & 0 & 1 & 2 & 5 \\
\hline
0 & 1.00 & 0.18 & 0.18 & 0.19 \\
1 & - & 1.00 & 0.73 & 0.64 \\
2 & - & - & 1.00 & 0.74 \\
5 & - & - & - & 1.00 \\
\midrule
\multicolumn{5}{c}{IAS}\\
\midrule
$h$ & 0 & 1 & 2 & 5 \\
\hline
0 & 1.00 & 0.43 & 0.31 & 0.28 \\
1 & - & 1.00 & 0.63 & 0.50 \\
2 & - & - & 1.00 & 0.72 \\
5 & - & - & - & 1.00 \\
\bottomrule
\end{tabular}
\caption{Cross-horizon ($h$) Pearson correlation coefficients for GPT at $T=1.5$ (upper panel), $T=0$ (middle panel), and IAS responses (lower panel). Sources: IAS and authors' calculations.}
\label{tab:time_corr}
\end{table}

\subsubsection{Model time trends}

\begin{figure}
\centering
\includegraphics[width=.7\textwidth]{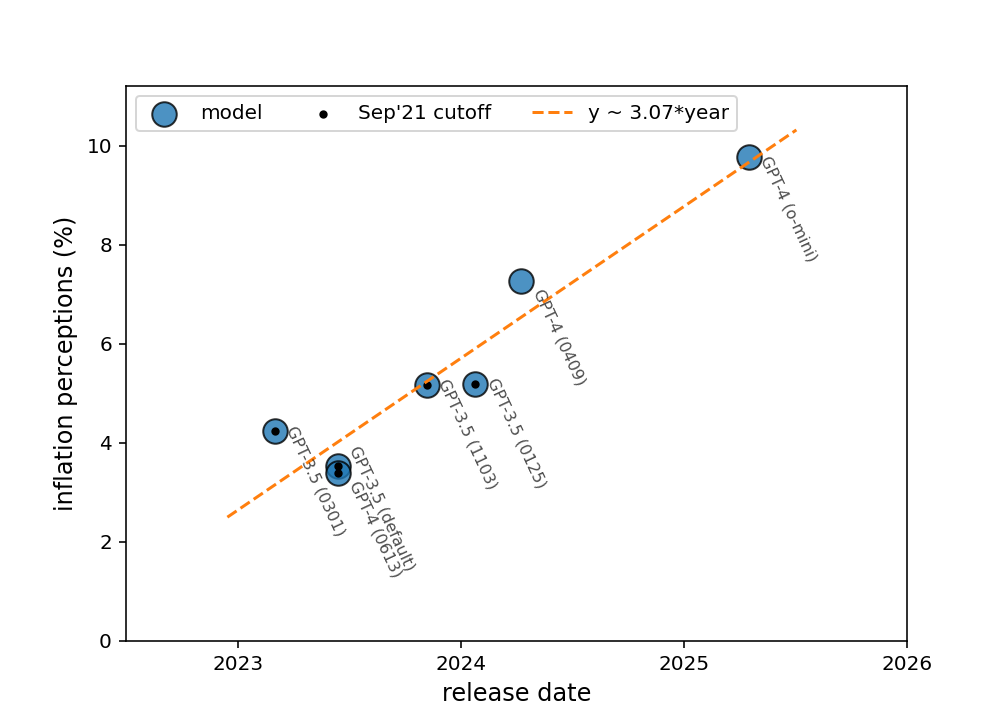}
\caption{Time trend of unconditioned inflation perceptions for different GPTs. Each circle is the simple mean of a subsample of the same 200 respondents from the main sample for 9 answer permutations for each model. Models with an official knowledge cut-off of September\ 2021 are marked with a dot. Sources: Authors' calculations.}
\label{fig:time_trends}
\end{figure}

Consumer price inflation and the ``cost of living crisis'' have been persistent topics of debate in the UK since the 2022 inflation spike\footnote{\url{https://commonslibrary.parliament.uk/research-briefings/cbp-9428/}}. That is, inflation has become a more salient topic in the public discourse. At the same time, LLM fine-tuning using reinforcement learning based on human user preferences has become common \citep{ouyang2022training}. However, such fine tuning can lead to an unpredictable level of information leakage from beyond the stated training cut-off. To test whether there may have been such leakages or salience of high inflation in ChatGPT's model family, we test unconditioned inflation perceptions for the main sample for different models, where we remove the economic conditioning information from the prompt only leaving demographics and the survey instructions. \\
The results of this exercise are summarised in Figure~\ref{fig:time_trends}, which plots average inflation perceptions against the release dates of the different models. We see that there is a clear time trend of an increase of about 3 p.p. unconditional inflation perceptions per year potentially creating large biases in later models as inflation numbers quickly fell after the spike.\\
Models with the same stated knowledge cut-off in September\ 2021 but later release dates seem to be affected by this trend suggesting the use of reinforcement learning and subsequent information leakage. Models released in early 2023 seem not to be affected by this, which highlights the value of the experimental setting for model validation and testing presented here.\\
This also highlights challenges for the use of closed-weight proprietary models like those from the ChatGPT family. The subsequent analyses will provide ways to either account for such ``offsets'' or to analyse the salience of different pieces of information to a model.

% >>> time trend

\subsection{Demographic drivers}
\label{sec:demo}

When driven by food and energy price growth, high inflation may be more concerning for lower income households, as they spend a larger proportion of their consumption basket on necessities like food and energy. By similar arguments, the inflation rate experienced by different demographic groups may vary.\\
Now, for a category $D_c$ we hypothesise the relative ordering of experienced inflation values $\mathcal{O}_c=\{\pi_{c_1}>\dots>\pi_{c_j}>\dots>\pi_{c_K}\}$ for $j\in\{1,\dots,K\}$ different classes, for example based on assumptions about their respective consumption baskets in a given economic environment. Taking $j=1$ as the reference class for each $D_c$, we can formulate the following hypothesis tests for jointly assessing $\mathcal{O}_c$ for categories $c\in\{1,\dots,C\}$ based on the response model
\begin{equation}
\label{eq:hypo}
y_{i,s}\,=\,b+\sum_{c=1}^{C} \sum_{j=2}^K \beta_{c_j} d_{ic_j} + \epsilon_i\,\quad\text{with}\quad \mathcal{H}_{cj}: \beta_{c_j}<0\;\text{for}\;j>1\,,
\end{equation}
where $s$ denotes the source of the $y_i$ (IAS or GPT in our case), $b$ is a constant, and $d_{ic_j}$ is a vector of dummies encoding the demographic profile of subject $i$. The coefficient vector $\beta=(\beta_{1_2},\dots,\beta_{C_K})'$ captures the joint relation between demographics and survey responses given a scenario. The interpretation of the elements of $\beta$ is intuitive, namely by how many percentage points (p.p.) inflation perceptions or expectations are on average higher or lower (depending on the sign of an element) if a subject belongs to a certain demographic group, i.e.\ $d_{ic_j}=1$.\\
An appealing property of \eqref{eq:hypo} is that it adjusts for a different location and scale of the response distribution $y_s$ via $b$ and the magnitude the $\beta$s, respectively. This means that, despite potentially poorer aggregate fits of a response distribution to a benchmark, we still can make inference about its drivers. This will be the case for the GPT ($T=0$) cases, which showed poorer aggregate matches to the IAS distribution but stronger correlations on the micro level (see Table~\ref{tab:time_profile}).\\
In the context of the main sample scenario in Table~\ref{tab:ec}, we hypothesise that the following reference groups within our demographic categories have experienced the highest levels of inflation: income: less than £9999 (lowest income), housing: council house, age: 16-24 (youngest), social class: working class, education: GCSEs but not A-levels (lowest formal education), region: Scotland. Consequently, when fitting \eqref{eq:hypo} we expect all components of $\beta$ to be {\it negative}.\\

\subsubsection{Inflation perceptions}
Table~\ref{tab:demo_reg} summarises the results for the IAS and again for GPT for $T=1.5$ and $T=0$ for inflation perceptions. We also include estimates of the actually experienced inflation by the different demographics categories at the time of the survey based on official statistics (ONS reference). Focusing on the $T=0$ case, we see that the majority of the coefficients are indeed negative and statistically highly significant indicating that GPT's responses are in line with the economic intuition guiding the choice of the reference classes, i.e.\ our hypotheses about which demographic groups may be more affected by the inflation spike around the time of the survey.\\
We can further validate this intuition and GPT's responses by comparing the $\beta$ estimates with the ONS reference. We indeed see that GPT responses are very much aligned with actual realisations for most categories when comparing directions: most entries are negative.\footnote{A notable exception is the coefficient of the ONS reference point for the lowest income group which is most likely is due to changes in the definition of lowest income group within the IAS survey itself over the years.} The exception to this are regional differences, where GPT thought they would be large, while there where almost none, perhaps because energy prices are regulated and determined mostly on the national and not regional level. \\
Comparing GPT responses to the IAS, we see that most coefficients are negative again. However, there is a major discrepancy between the two for age. While GPT thinks that there is a mostly negative and marginally increasing effect with age, IAS respondents believe that the effect is clearly positive and strongly increasing. A possible effect is the observed cohort effect of inflation perceptions \cite{MalmendierNagel2013}: older cohorts may have experienced more ``scarring'' by periods of high inflation and, as a consequence, have on average more pessimistic views on inflation in an heightened inflation environment.\\

% >>> summary table with (humans, T=0, T=1.5)
\begin{table}[htbp]
\centering
\begin{tabular}{lccc|c}
\toprule
variable & IAS & GPT (T=1.5) & GPT (T=0) & ONS ref. \\
\midrule
income (£10000-£19999) & -0.25\phantom{$^{***}$} & \phantom{-}0.91$^{**}$\phantom{$^{*}$} & -0.16\phantom{$^{***}$} & 0.16 \\
income (£20000-£34999) & -0.29\phantom{$^{***}$} & \phantom{-}0.53\phantom{$^{***}$} & -1.10$^{***}$ & -0.20 \\
income (£35000-£44999) & -0.45\phantom{$^{***}$} & \phantom{-}0.40\phantom{$^{***}$} & -1.35$^{***}$ & -0.35 \\
income (>£45000) & -0.31\phantom{$^{***}$} & \phantom{-}0.54\phantom{$^{***}$} & -0.87$^{***}$ & -0.68 \\
\midrule
housing (private renter) & \phantom{-}0.36\phantom{$^{***}$} & -0.61$^{*}$\phantom{$^{**}$} & -0.36\phantom{$^{***}$} & -0.27 \\
housing (owner w/mortgage) & -0.10\phantom{$^{***}$} & -0.70$^{**}$\phantom{$^{*}$} & -1.23$^{***}$ & -1.54 \\
housing (outright owner) & -0.59$^{*}$\phantom{$^{**}$} & -0.70$^{*}$\phantom{$^{**}$} & -1.03$^{***}$ & -1.39 \\
\midrule
age (25-34) & \phantom{-}1.47$^{***}$ & -0.66$^{**}$\phantom{$^{*}$} & -0.99$^{***}$ & -0.12 \\
age (35-44) & \phantom{-}2.53$^{***}$ & -0.88$^{***}$ & -0.87$^{***}$ & -0.28 \\
age (45-54) & \phantom{-}3.96$^{***}$ & -0.98$^{***}$ & -0.80$^{***}$ & -0.46 \\
age (55-64) & \phantom{-}5.02$^{***}$ & -0.61$^{*}$\phantom{$^{**}$} & -0.39\phantom{$^{***}$} & -0.58 \\
age (65-75) & \phantom{-}4.91$^{***}$ & -0.18\phantom{$^{***}$} & \phantom{-}0.53$^{*}$\phantom{$^{**}$} & -0.39 \\
\midrule
class (skilled working) & -0.87$^{***}$ & -0.43\phantom{$^{***}$} & -0.37\phantom{$^{***}$} & -0.14 \\
class (lower middle) & -0.50$^{*}$\phantom{$^{**}$} & -0.34\phantom{$^{***}$} & -0.34\phantom{$^{***}$} & -0.52 \\
class (upper middle) & -0.75$^{**}$\phantom{$^{*}$} & -0.83$^{**}$\phantom{$^{*}$} & -0.74$^{***}$ & -0.62 \\
class (pensioner) & -1.39$^{**}$\phantom{$^{*}$} & -2.01$^{***}$ & -1.78$^{***}$ & -0.28 \\
\midrule
education (A-levels) & \phantom{-}0.45$^{*}$\phantom{$^{**}$} & -0.41\phantom{$^{***}$} & -0.16\phantom{$^{***}$} & 0.06 \\
education (degree) & \phantom{-}0.18\phantom{$^{***}$} & \phantom{-}0.11\phantom{$^{***}$} & \phantom{-}0.01\phantom{$^{***}$} & 0.08 \\
\midrule
region (North or Northern Ireland) & \phantom{-}0.30\phantom{$^{***}$} & \phantom{-}0.25\phantom{$^{***}$} & -0.14\phantom{$^{***}$} & 0.09 \\
region (West of England or Wales) & -0.07\phantom{$^{***}$} & \phantom{-}0.31\phantom{$^{***}$} & -0.52$^{*}$\phantom{$^{**}$} & 0.02 \\
region (Midlands) & \phantom{-}0.01\phantom{$^{***}$} & \phantom{-}0.61$^{*}$\phantom{$^{**}$} & -0.53$^{**}$\phantom{$^{*}$} & 0.03 \\
region (South East) & -0.39\phantom{$^{***}$} & \phantom{-}0.60$^{*}$\phantom{$^{**}$} & -0.80$^{***}$ & 0.30 \\
\midrule
constant ($b$) & \phantom{-}6.41$^{***}$ & \phantom{-}10.15$^{***}$ & \phantom{-}13.72$^{***}$ & - \\
%\hline
n obs. & 3808 & 3808 & 3808 & - \\
$R^2$ (adj.) & 0.11 & 0.01 & 0.06 & - \\
\bottomrule
\end{tabular}
\caption{Summary table for weighted least square regressions of individual survey responses (inflation perceptions) for IAS, GPT (T=1.5), and GPT (T=0) on demographic dummies against base classes (housing: council house, income: income of <£9999, age: 16-24, social class: working class, education: GCSEs but not A-levels, region: Scotland). {*}{*}{*} indicates significance at the 1\% level, {*}{*} at the 5\%
level, and {*} at the 10\% level. Robust standard errors have been used. The right-most column states ONS reference values for February 2023 inflation as the differences in percentage points against the corresponding base class. Sources: IAS, ONS, and authors' calculations.}
\label{tab:demo_reg}
\end{table}
%
% >>> regression plots (housing, income, age)
\begin{figure}
\centering
\includegraphics[width=.95\textwidth]{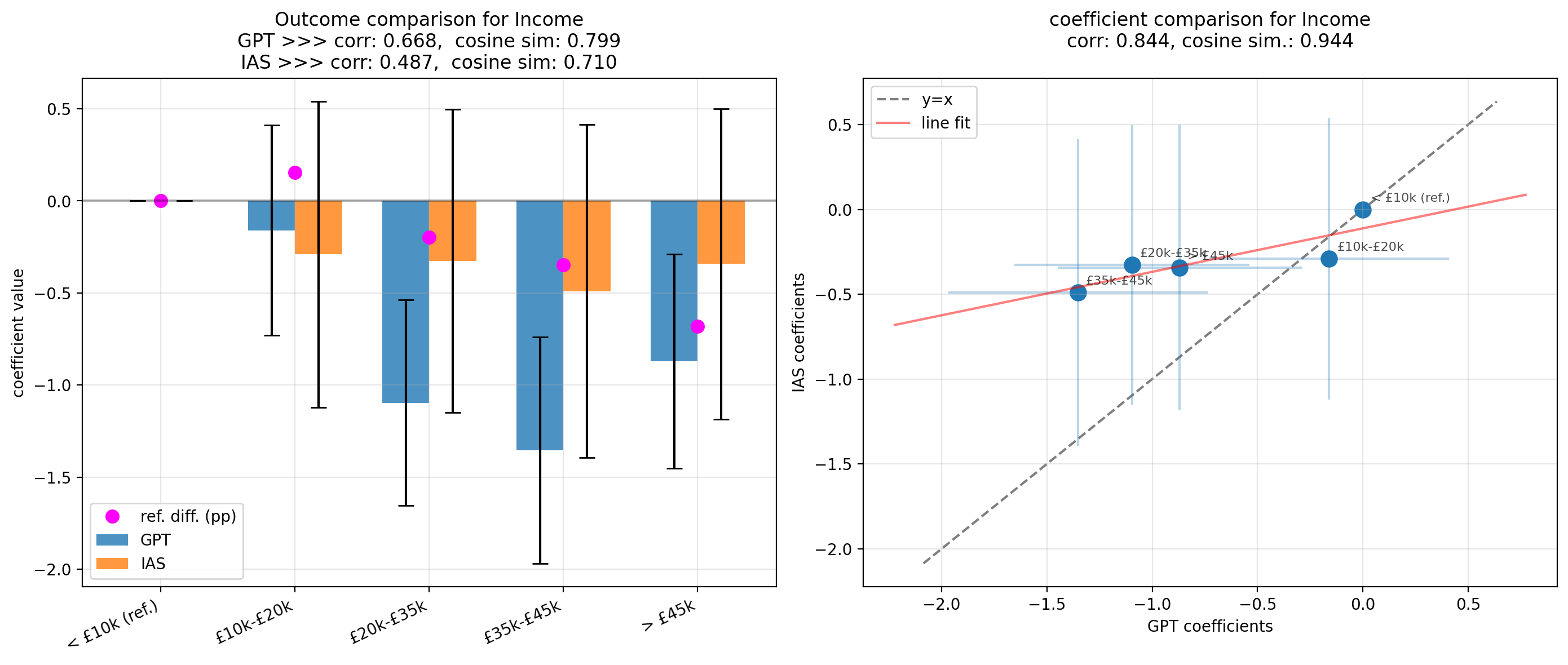}\\
\includegraphics[width=.95\textwidth]{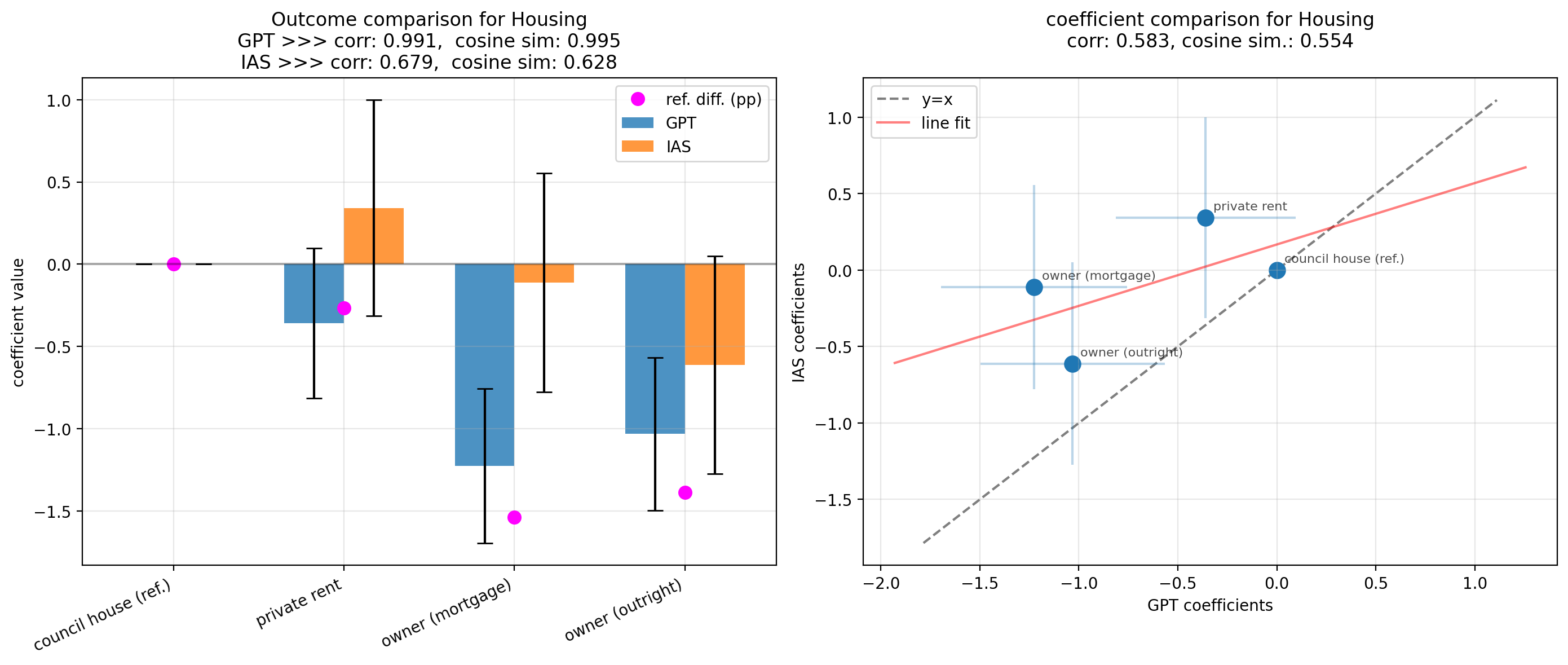}\\
\includegraphics[width=.95\textwidth]{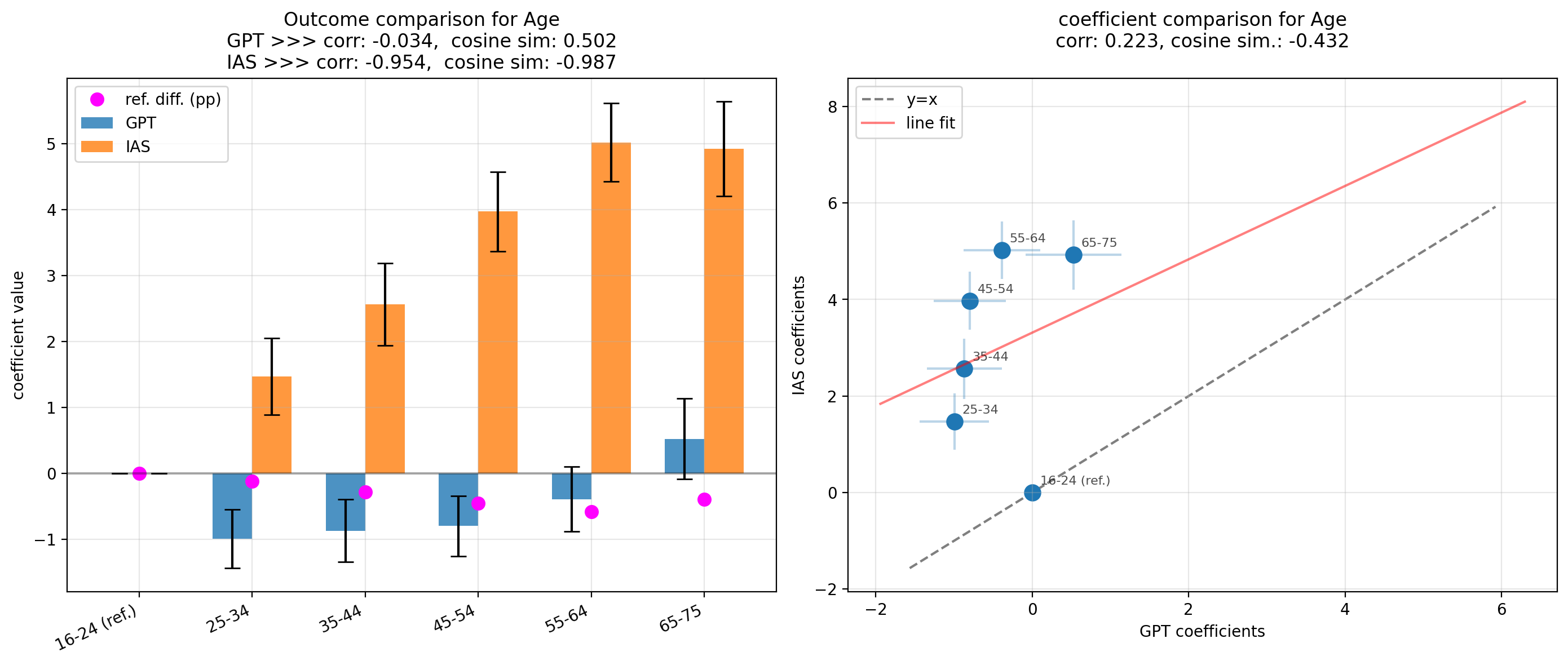}
\caption{Graphical summaries of demographic regressions of GPT ($T=0$) inflation perceptions for income (upper panel), housing tenure (middle panel), and age (lower panel). LHS: three-way comparison between IAS (sand), GPT (blue), and ONS (magenta). RHS: scatter plot of IAS coefficients versus GPT coefficients. Best-fit line given in red. Error bars indicate 95\% confidence intervals throughout. Summary stats are Pearson correlation coefficient (corr) and cosine similarity. Sources: IAS, ONS, authors' calculations.}
\label{fig:demo_reg}
\end{figure}

The demographic regression results are graphically summarised for the first three categories of Table~\ref{tab:demo_reg} in Figure~\ref{fig:demo_reg}. The left-hand side panels compare IAS and GPT coefficients against ONS out-turns as measured by the difference against the reference class in p.p.. The right-hand side panels compare GPT and IAS coefficients against each other on a scatter plot: the more dots align along the diagonal as summarised by the fit line, the more similar human and GPT responses are alongside this demographics dimension.\\
We see that there is reasonable agreement between GPT, IAS, and ONS out-turns for income and especially for housing. GPT's response patterns closely match that of the outcome. This is interesting and encouraging as there are various channels how different housing tenures can affect realised inflation values suggesting that GPT can be useful in modelling complex and potentially unknown relationships.\\
However, our analysis also suggests caution when looking at the results for income in the bottom part of Figure~\ref{fig:demo_reg}. Here the three-way comparison is far less aligned. As pointed out before, GPT and IAS response patterns are mostly opposed to each other with some similarity in trends with increasing age. The alignment with the out-turn is weak at best.\footnote{The corresponding figure is given in the Appendix for $T=1.5$, where see that micro-level relations generally weaken with higher temperature.}\\
This example raises the important question how to interpret and use GPT as a modelling tool: should we prefer GPT to be more closely aligned with the human sample we emulate or with the survey quantities we elicit? We will address such concerns in the discussion at the end of the paper. %The answer to this question may not always be straightforward, and it may depend on the goal of the analysis and how its outputs are being used. For example, if the goal is to infer statistics which are otherwise hard to measure, trusting GPT's output may be less problematic after validation. However, using GPT can be more problematic in cases where its outputs are being used to inform decision relating to groups with whom it does not align well. The techniques we present here offer approaches for making these assessments. 
%
% >>> summary table with (humans, T=0, T=1.5) [EXPECTATIONS]
\begin{table}[htbp]
\centering
\begin{tabular}{lccc|c}
\toprule
variable & IAS & GPT (T=1.5) & GPT (T=0) & ONS ref. \\
\midrule %\phantom{$-$}
income (£10000-£19999) & -0.54\phantom{$^{***}$} & \phantom{-}0.21\phantom{$^{***}$} & -0.41$^{**}$\phantom{$^{*}$} & \phantom{-}0.03 \\
income (£20000-£34999) & -0.81$^{*}$\phantom{$^{**}$} & -0.19\phantom{$^{***}$} & -0.75$^{***}$ & -0.05 \\
income (£35000-£44999) & -1.14$^{***}$ & -0.11\phantom{$^{***}$} & -0.83$^{***}$ & -0.01 \\
income (>£45000) & -1.18$^{***}$ & -0.55$^{*}$\phantom{$^{**}$} & -0.85$^{***}$ & -0.03 \\
\midrule
housing (private renter) & -0.15\phantom{$^{***}$} & -0.98$^{***}$ & -0.79$^{***}$ & \phantom{-}0.21 \\
housing (owner w/mortgage) & -0.29\phantom{$^{***}$} & -0.88$^{***}$ & -0.87$^{***}$ & -0.16 \\
housing (outright owner) & \phantom{-}0.49\phantom{$^{***}$} & -0.44$^{**}$\phantom{$^{*}$} & -0.41$^{***}$ & \phantom{-}0.33 \\
\midrule
age (25-34) & \phantom{-}2.35$^{***}$ & -0.07\phantom{$^{***}$} & -0.14\phantom{$^{***}$} & \phantom{-}0.10 \\
age (35-44) & \phantom{-}0.77$^{***}$ & -0.27\phantom{$^{***}$} & -0.34$^{***}$ & \phantom{-}0.24 \\
age (45-54) & \phantom{-}2.09$^{***}$ & \phantom{-}0.50$^{**}$\phantom{$^{*}$} & -0.05\phantom{$^{***}$} & \phantom{-}0.26 \\
age (55-64) & \phantom{-}1.56$^{***}$ & \phantom{-}1.05$^{***}$ & \phantom{-}0.22\phantom{$^{***}$} &\phantom{-}0.28 \\
age (65-75) & \phantom{-}1.16$^{***}$ & -0.04\phantom{$^{***}$} & -0.22\phantom{$^{***}$} & \phantom{-}0.37 \\
\midrule
class (skilled working) & -0.90$^{***}$ & -0.07\phantom{$^{***}$} & -0.20\phantom{$^{***}$} & \phantom{-}0.10 \\
class (lower middle) & -0.84$^{***}$ & -0.46$^{**}$\phantom{$^{*}$} & -0.67$^{***}$ & \phantom{-}0.01 \\
class (upper middle) & -0.99$^{***}$ & -0.78$^{***}$ & -1.18$^{***}$ & -0.07 \\
class (pensioner) & -0.92\phantom{$^{***}$} & -2.33$^{***}$ & -1.21$^{***}$ & \phantom{-}0.18 \\
\midrule
education (A-levels) & \phantom{-}0.33\phantom{$^{***}$} & -0.06\phantom{$^{***}$} & -0.05\phantom{$^{***}$} & \phantom{-}0.05 \\
education (degree) & \phantom{-}0.50$^{**}$\phantom{$^{*}$} & -0.58$^{***}$ & -0.21$^{**}$\phantom{$^{*}$} & \phantom{-}0.01 \\
\midrule
region (North or Northern Ireland) & -0.19\phantom{$^{***}$} & -0.51$^{**}$\phantom{$^{*}$} & \phantom{-}0.20\phantom{$^{***}$} & \phantom{-}0.08 \\
region (West of England or Wales) & -0.51\phantom{$^{***}$} & -0.47$^{*}$\phantom{$^{**}$} & -0.09\phantom{$^{***}$} & -0.01 \\
region (Midlands) & -0.39\phantom{$^{***}$} & -0.51$^{**}$\phantom{$^{*}$} & -0.10\phantom{$^{***}$} & \phantom{-}0.06 \\
region (South East) & -0.46\phantom{$^{***}$} & -0.26\phantom{$^{***}$} & \phantom{-}0.04\phantom{$^{***}$} & \phantom{-}0.11 \\
\midrule
constant ($b$) & \phantom{-}5.43$^{***}$ & \phantom{-}8.02$^{***}$ & \phantom{-}8.37$^{***}$ & - \\
%\hline
n obs. & 3636 & 3572 & 3636 & - \\
$R^2$ (adj.) & 0.03 & 0.09 & 0.14 & - \\
\bottomrule
\end{tabular}
\caption{Summary table for weighted least square regressions of individual survey responses (inflation expectations) for IAS, GPT (T=1.5), and GPT (T=0) on demographic dummies against base classes (housing: council house, income: income of <£9999, age: 16-24, social class: working class, education: GCSEs but not A-levels, region: Scotland). {*}{*}{*} indicates significance at the 1\% level, {*}{*} at the 5\%
level, and {*} at the 10\% level. Robust standard errors have been used. The right-most column states ONS reference values for February 2023 inflation as the differences in percentage points against the corresponding base class. Sources: IAS, ONS, and authors' calculations.}
\label{tab:demo_reg_exp}
\end{table}
%
% >>> regression plots (housing, income, age) (EXPECTATIONS)
\begin{figure}
\centering
\includegraphics[width=.95\textwidth]{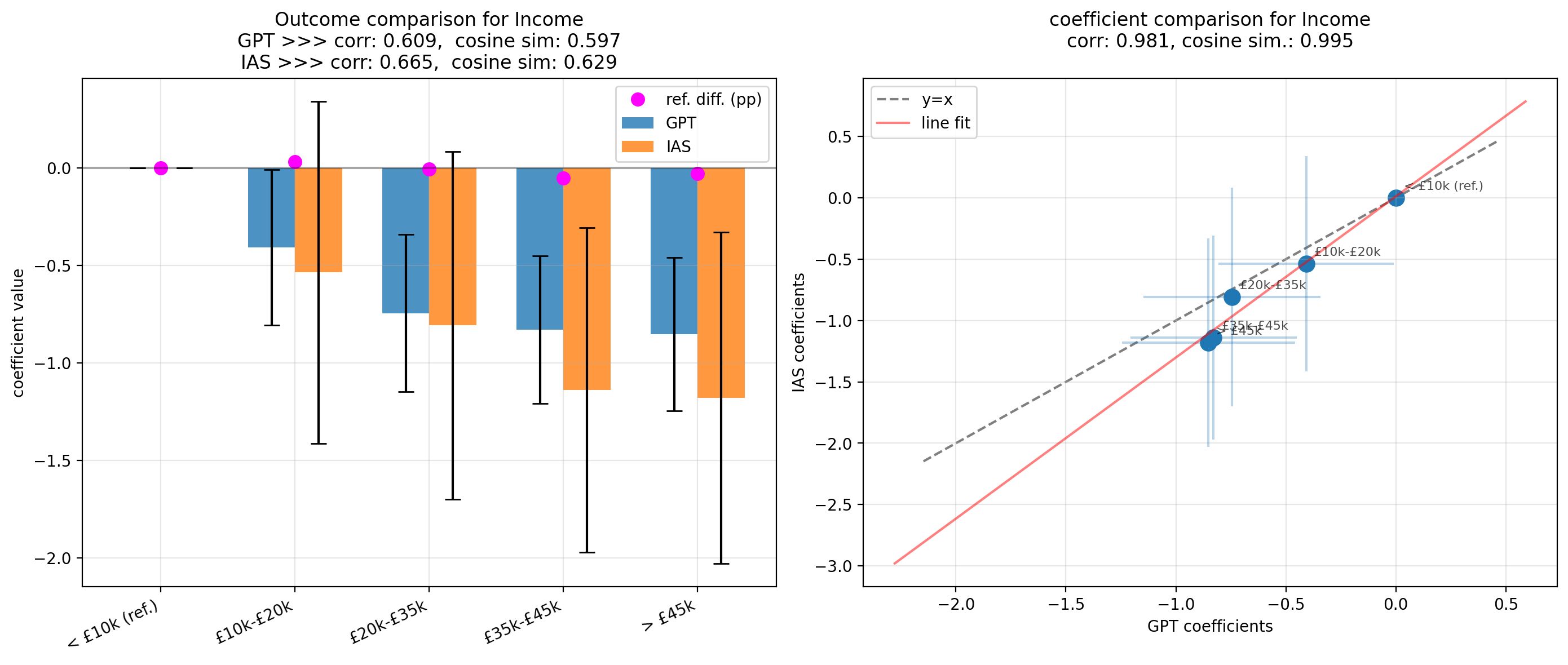}\\
\includegraphics[width=.95\textwidth]{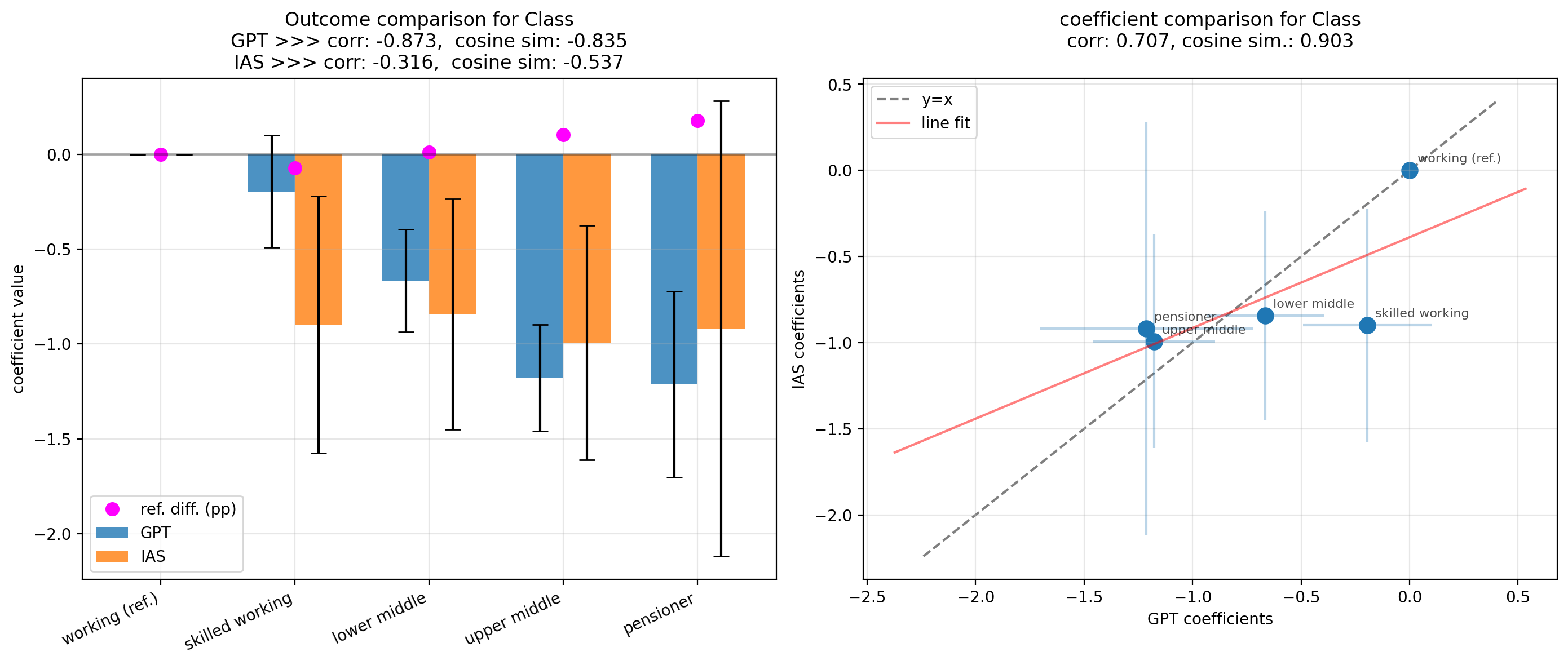}\\
\includegraphics[width=.95\textwidth]{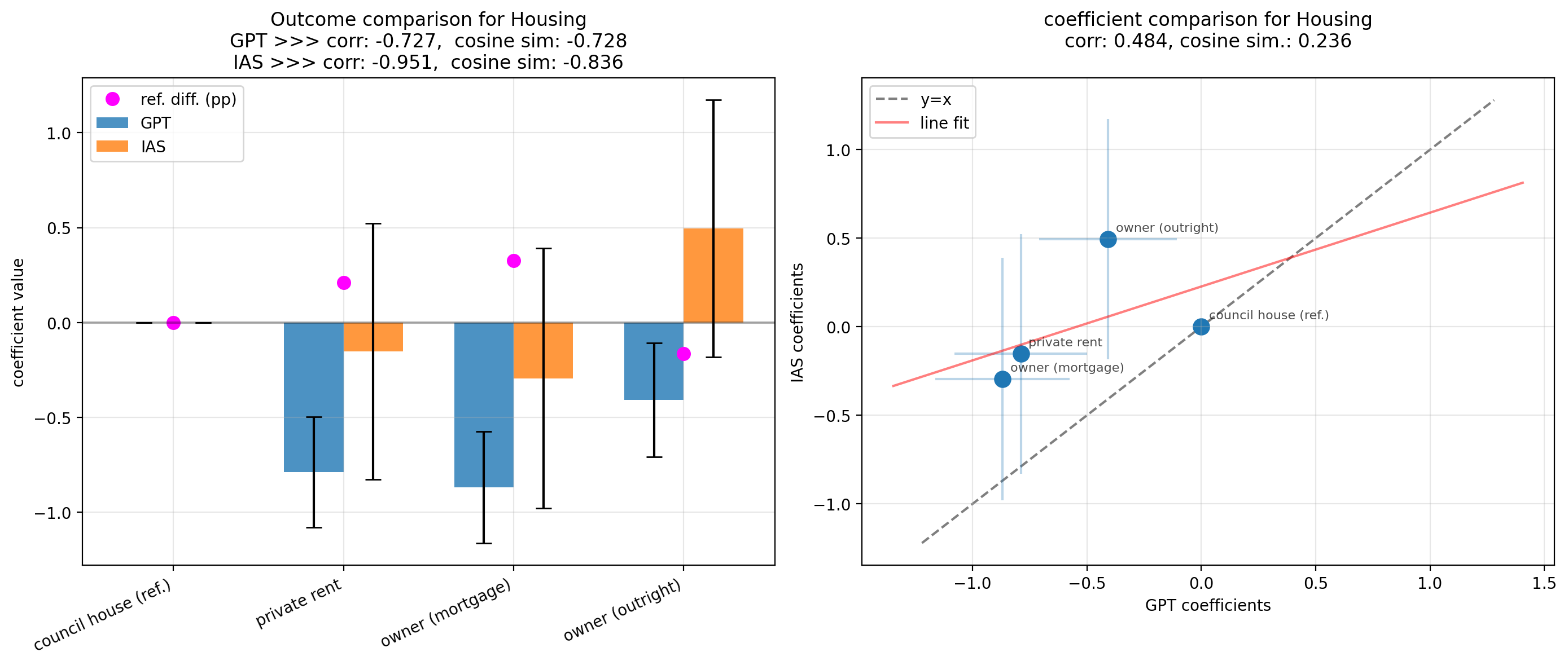}
\caption{Graphical summaries of demographic regressions of GPT ($T=0$) inflation expectations for income (upper panel), social class (middle panel), and housing tenure (lower panel). LHS: three-way comparison between IAS (sand), GPT (blue), and ONS (magenta). RHS: scatter plot of IAS coefficients versus GPT coefficients. Best-fit line given in red. Error bars indicate 95\% confidence intervals throughout. Summary stats are Pearson correlation coefficient (corr) and cosine similarity. Sources: IAS, ONS, authors' calculations.}
\label{fig:demo_reg_exp}
\end{figure}

\subsubsection{Inflation expectations}

\begin{figure}
\centering
\label{fig:demo_reg_app}
\includegraphics[width=.95\textwidth]{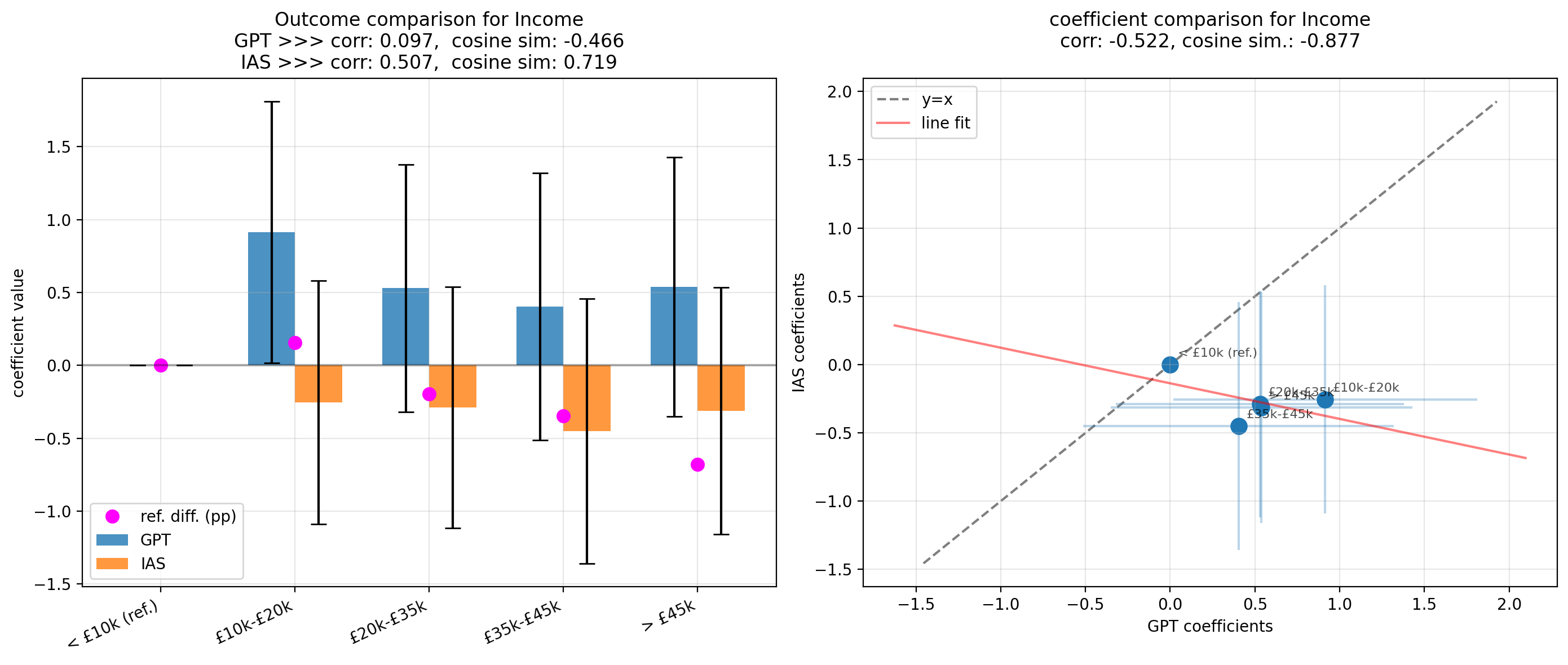}\\
\includegraphics[width=.95\textwidth]{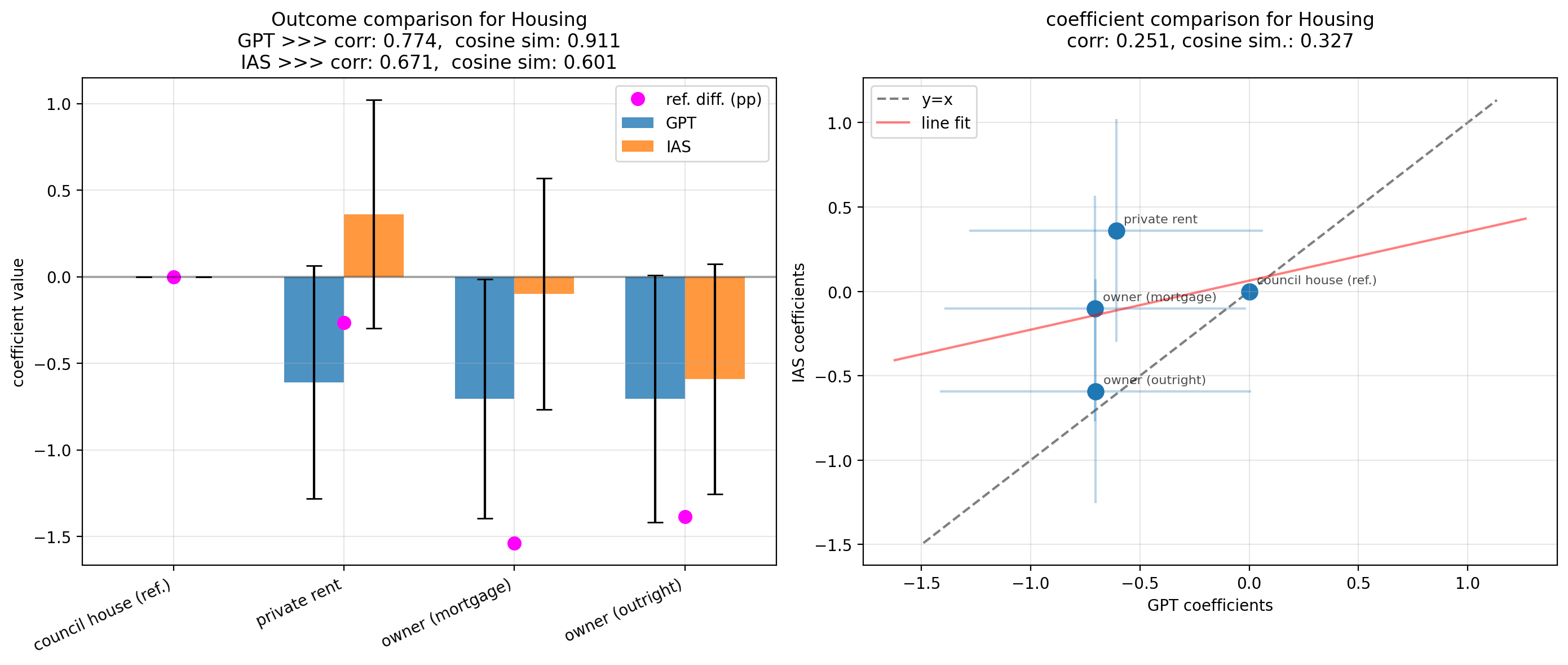}\\
\includegraphics[width=.95\textwidth]{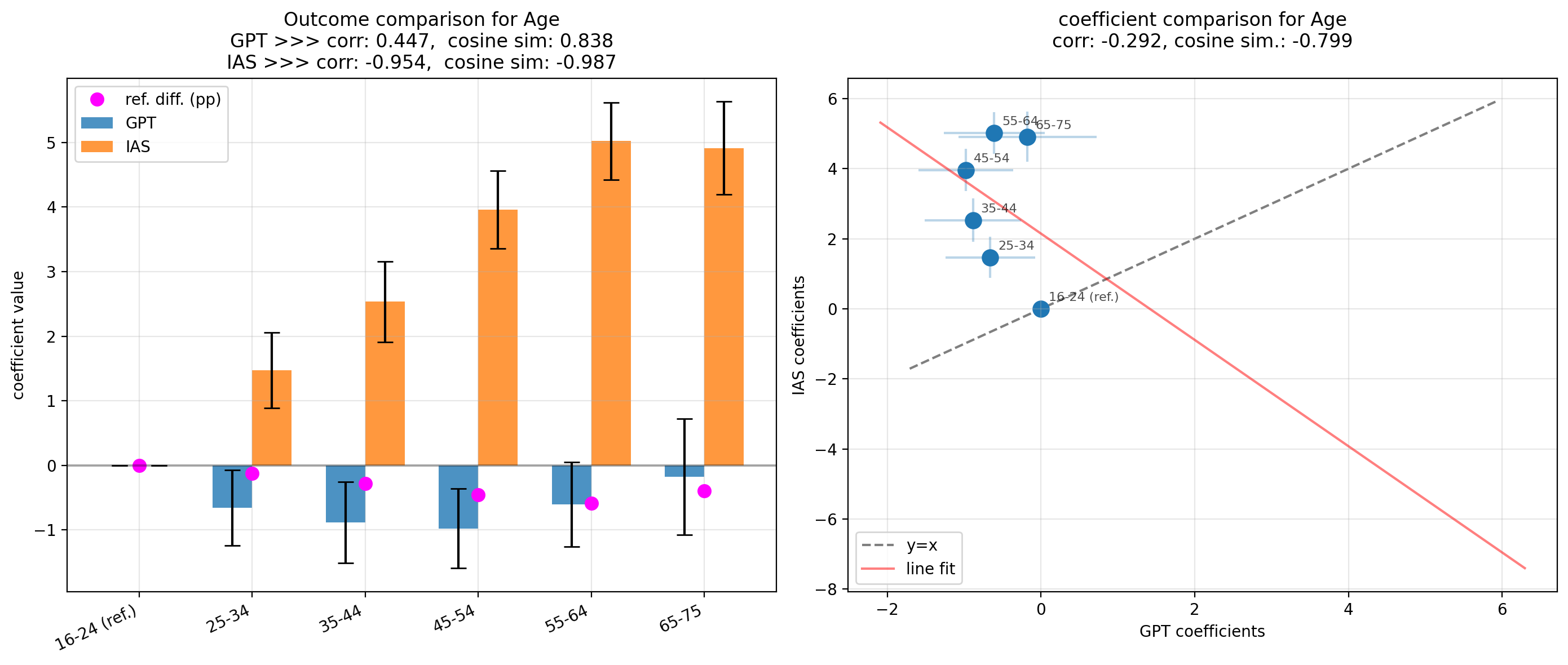}
\caption{Graphical summaries of demographic regressions of GPT ($T=1.5$) inflation perceptions for income (upper panel), housing tenure (middle panel), and age (lower panel). LHS: three-way comparison between IAS (sand), GPT (blue), and ONS (magenta). RHS: scatter plot of IAS coefficient versus GPT coefficients. Best-fit line given in red. Error bars indicate 95\% confidence intervals throughout. Summary stats are Pearson correlation coefficient (corr) and cosine similarity. Sources: IAS, authors' calculations.}
\end{figure}

Perceptions are easier to compare with realised values as there is less noise affecting the analysis between eliciting survey responses and measuring the reference quantities. However, many economic concepts relate to expectations at longer horizons, often at one quarter or one year out. We therefore repeat the exercise from the previous subsection for inflation expectations at the one-year horizon. The results are summarised in Table~\ref{tab:demo_reg_exp}, where we will focus on the $T=0$ case. \\
Looking at explanatory power ($R^2$), it is interesting to observe that GPT now shows more alignment with demographics on average compared to the IAS. The comparison with realised values (last column of Table~\ref{tab:demo_reg_exp}) is less interesting now because aggregate inflation decreased markedly from 9.2\% (February 2023) to 3.8\% (February 2024), and the variation within demographic groups is much smaller compared to when inflation peaked.\\
The patterns of signs persist between inflation perceptions and one-year expectations: most signs are negative, again with exception for age for the IAS. Looking at individual demographics in Figure~\ref{fig:demo_reg_app}, we see that there is good alignment between GPT and the IAS for income, housing, and social class. This congruence is particularly strong for the income distribution, a main quantity of interest when investigating heterogeneity in economic analyses.

\subsection{Economic drivers}
\label{sec:econ}

We investigate how the details of the economic conditioning affect GPT's ($T=0$) predictions for inflation perceptions in the main sample. 

\subsubsection{Marginal sensitivity}
\label{sec:marginal}

Aggregate inflation scales approximately linearly with inflation of the subcomponents of the indeces.\footnote{Approximately because the basket-weighted sum of component price growth does not need to be the same as overall index growth but will be a good approximation in most cases.} As such, it is an interesting question whether GPT's inflation perceptions scale linearly with the different components of the economic conditioning information treatment. To do so, we track the na\"ive treatment effect \eqref{eq:energy_naive} for food \& restaurants, energy, and the main other component with the offset being the respective aggregate prediction value at the zero input for that component.\footnote{For example, we subtract $MN(g(t=(0,0,\bar{t}_e,\bar{t}_o);x_i);w)$ for the food \& restaurants component.} In each case, we vary the respective component covering the historically observed values including the inflation spike in 2023Q1.\footnote{For food \& restaurants, we varied food price inflation keeping the ratio between this and restaurants \& cafes inflation constant at the value of the main scenario.}\\

% >>> marginal plots
\begin{figure}
\centering
\includegraphics[width=.6\textwidth]{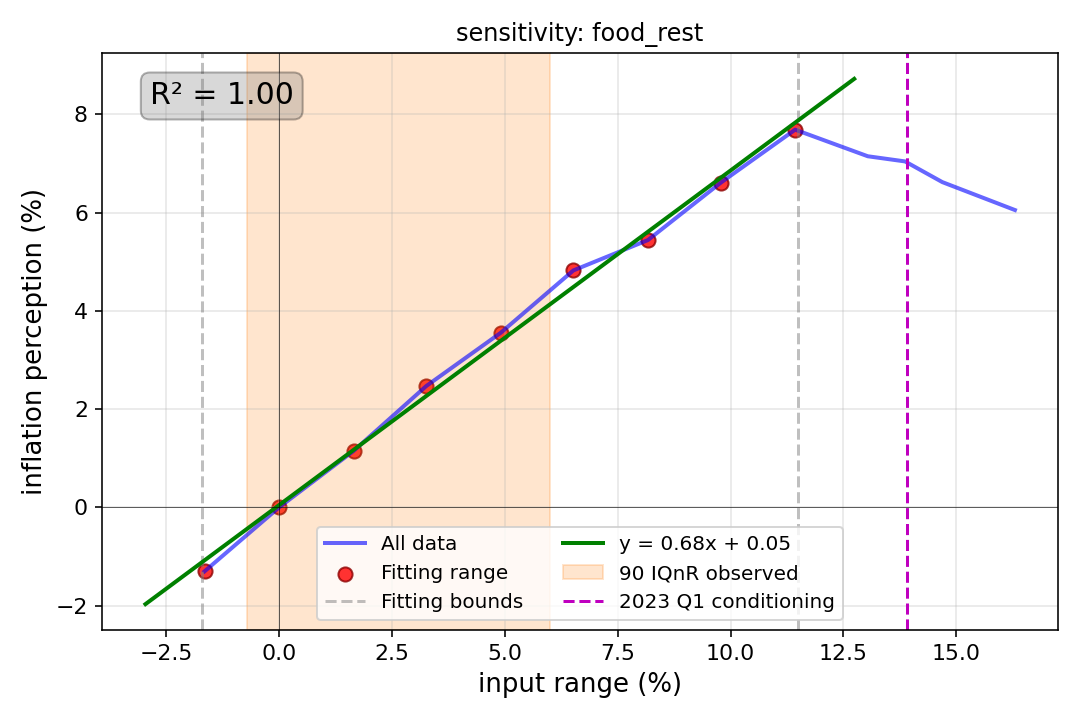}\\
\includegraphics[width=.6\textwidth]{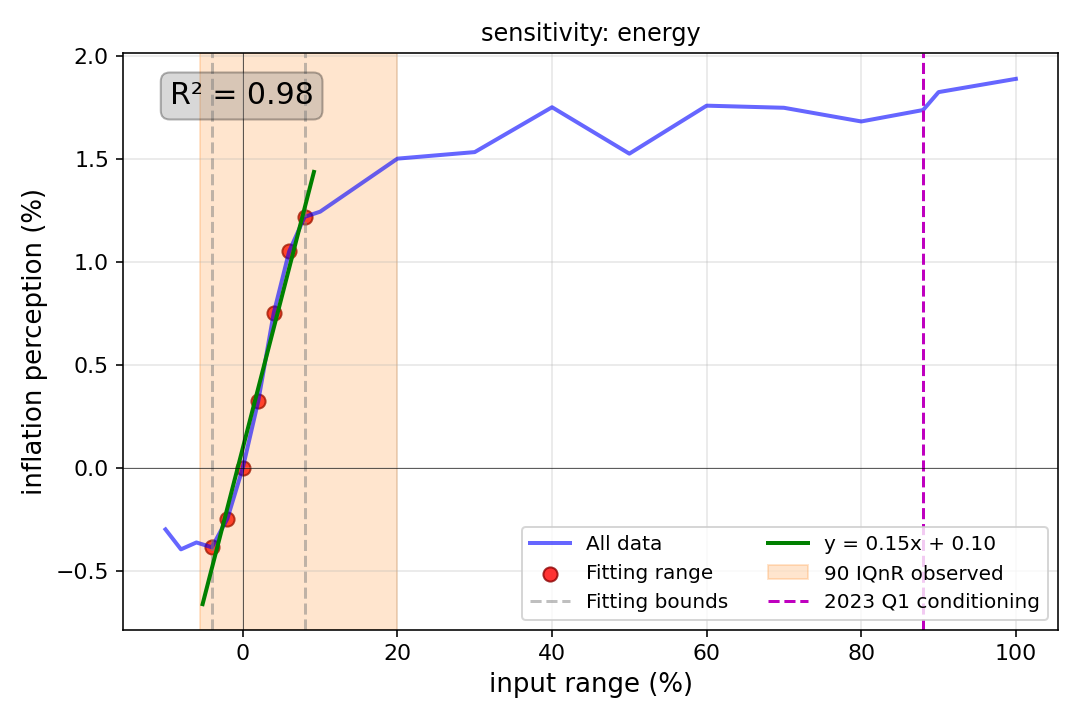}\\
\includegraphics[width=.6\textwidth]{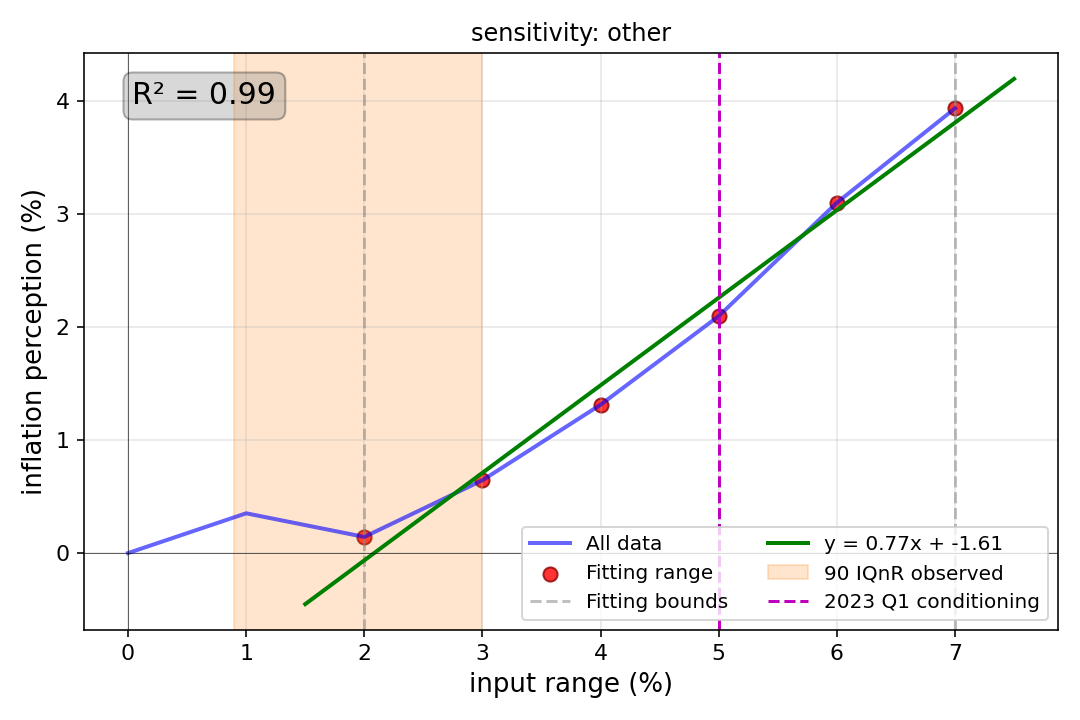}
\caption{Sensitivity analysis of GPT ($T=0$) inflation perceptions for varying individual index components across their respective observed ranges: food \& restaurants (upper panel, x-axis is index-weighted mean of both components), energy (middle panel), and the remaining other component. The shaded area covers the inter-90 percentile range of each component pre-Oct-2021. The purple dashed lined marks the economic conditions used in the main experiment. Sources: IAS, ONS, authors' calculations.}
\label{fig:sensitivity}
\end{figure}

The results of this exercise are shown in Figure~\ref{fig:sensitivity}. We again make several observations. First, GPT's aggregate perceptions indeed scale linearly with subcomponent inflation over large parts of the input space. However, this scaling does not consistently cover the historically observed input ranges (shaded areas) for some components (energy and other), but may extend far beyond them (food and other). GPT's behaviour is particularly puzzling for the main other component. It is mostly insensitive to this component between zero and two percent inflation but then scales linearly far beyond historically observed values.\\
Second, the slopes within the linear response ranges of GPT for the different components can be directly compared to the corresponding basket weights. This comparison is shown in the upper part of Table~\ref{tab:marginal}, where the ratio between the slope and the weight measures whether GPT over- or under-reacts to a component corresponding to a value bigger or smaller than one, respectively. GPT strongly over-reacts to information from the food \& restaurant and the energy components. This behaviour may be said to be similar to that of humans, since it is known that they over-rely on salient components such as groceries to make judgments about inflation \cite{DAcunto2021}. Interestingly, GPT's responsiveness to the main other component is very much proportional to its share in the basket beyond the 2\% point in the input range, going far beyond the historically observed range. This suggest that GPT can partially be used to make realistic inference about inflation once corrections for the observed bias at the origin are included.\\
Lastly, the actual and real scenario from the main sample is rather extreme. The vertical purple lines which mark the economic conditioning are far beyond the historically observed data ranges and also outside GPT's linear scaling range for food \& restaurants and energy. It is somewhat interesting that GPT fails to extrapolate monotonically with respect to food \& restaurants inflation. As such, our analysis sketches out the boundaries within which GPT's responses may be judged consistent but also highlights that GPT does not seem to have a complete internal model of consumer price inflation.

% >>> marginal summary table
\begin{table}[htbp]
\centering
\begin{tabular}{lccc|c}
\toprule
component & food\_rest & energy & other & sum \\
\midrule
slope  & 0.68 & 0.15 & 0.77 &  1.60\\
weight & 0.17 & 0.04 & 0.79 &  1.00\\
ratio  & 4.01 & 3.55 & 0.98 & (1.60)\\
\midrule
na\"ive & 7.04 & 1.74 & 2.10 & 11.87\\
Shapley & 6.18 & 2.79 & 1.76 & 11.73\\
\bottomrule
\end{tabular}
\caption{Summary of sensitivity analysis of GPT ($T=0$) inflation perceptions with respect to different sub-components (columns). Upper part: regression analysis from Fig.\ref{fig:sensitivity}. Lower part: contributions to main scenario treating components with independent (na\"ive) or possible interacting (Shapley) treatment component. The right-most columns provides the sum of columns except for the ratio. The mean GPT ($T=0$) inflation perception is 11.80. Sources: IAS, ONS, authors' calculations.}\label{tab:marginal}
\end{table}

\subsubsection{Shapley decomposition}
\label{sec:shapley}

In each panel of Figure~\ref{fig:sensitivity}, the value on the vertical axis where the 2023Q1 conditioning line intersects with the GPT response curve measures the na\"ive treatment effect \eqref{eq:energy_naive} for that component. As discussed, this decomposition does not take into account possible interactions between the treatment components. These are captured by the Shapley components \eqref{eq:shap}. A comparison between the two indicates how important such interactions are. Both are shown in the lower part of Table~\ref{tab:marginal}. The Shapley decomposition measures all components against the zero baseline, i.e.\ where non-active component values are set to zero.\\
The ranking of components in the main scenario is the same. However, we do see sizeable changes for the importance of the energy treatment, which increases by more than 50\% in absolute terms for the Shapley case. This suggests that GPT's inflation perceptions are particularly elevated due to jointly high values of food and energy price inflation. This is confirmed by considering a scenario with only these two components at non-zero values as in the main scenario. The effect for this scenario is 11.09\% of aggregate GPT inflation perceptions. The remainder compared to the full main scenario is accounted for by the other component and its interactions with the other two components.\footnote{The fact that neither the na\"ive nor the Shapley decomposition with zero values for inactive components perfectly sums up to the aggregate total of inflation perceptions in the main scenario suggests the existence of additional, though minor, inconsistencies in GPT's inflation attitudes.}\\
%

% >>> Shapley decomposition
\begin{figure}
\centering
\includegraphics[width=.7\textwidth]{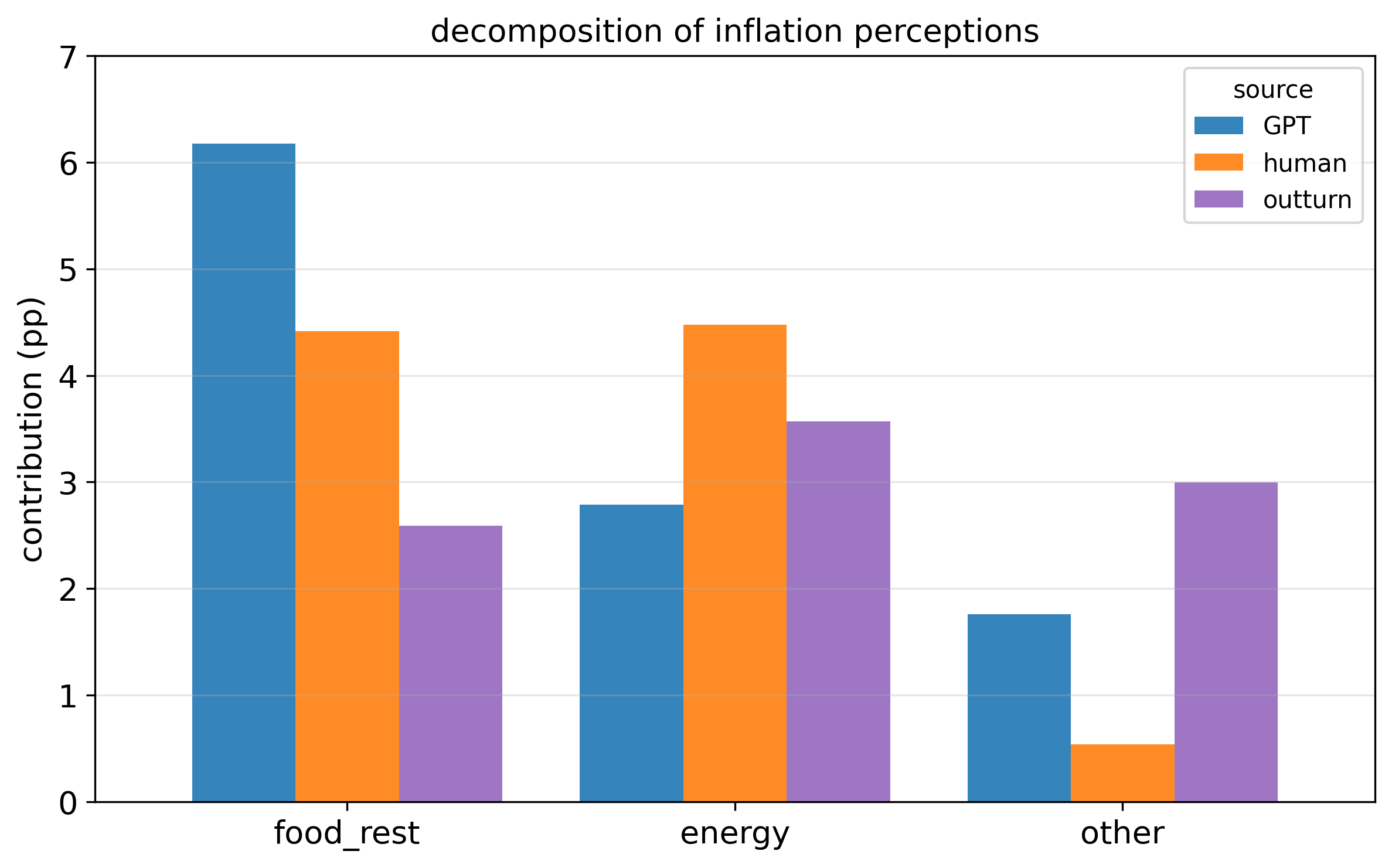}
\caption{Component decomposition for the main scenario for GPT ($T=0$): Shapley values against zero reference value (blue; $\tau^S_0=0.07$); IAS: regression model (sand), and index weighted sub-component inflation (purple).  Sources: IAS, ONS, authors' calculations.}
\label{fig:shapley}
\end{figure}

The relevance of a component for aggregate inflation is the product of its magnitude and weight. For instance, energy inflation was very ran very high in the main scenario while its weight is moderate at about 4\% in the consumption basket. By comparing GPT aggregate components with realised contributions we can assess whether it over- or under-weights individual components for a given scenario independent of its response function. This comparison is shown for the main scenario in Figure~\ref{fig:shapley}. In line with previous results, GPT heavily over-weights food \& restaurant inflation and under-weights the main other component. Its perception of the relevance of energy price inflation may be said to be roughly in line with its actual importance.\\
Finally, an advantage of the GPT decompositions approaches presented here is that we can make such inference in the first place. To get to something comparable for the actual survey, we need to build a model of how human respondents react to economic conditions. A possible approach to do so is to regress aggregate survey responses on the economic conditions at the time of the survey across a certain time period. We can then measure the  importance of each component for a given scenario for this model by multiplying the coefficient values with the respective scenario input values, which corresponds to the Shapley value of a variable in a linear model.\\
Modelling results for different fit ranges are shown in Table~\ref{tab:compo_coeffs}. We see that such relations can be rather unstable over time. Only the relevance of the food \& restaurant component is stable over time.\footnote{All coefficients turn insignificant for a short modelling period of only five years.} The Shapley decomposition corresponding to the first row of Table~\ref{tab:compo_coeffs} is shown in orange in Figure~\ref{fig:shapley}. A possibly robust statement we can make from this comparison is that humans and GPT overreact to food \& restaurant inflation in line with previous findings providing some trust in the use of GPT in this context.

% >>> EC regression summary table
\begin{table}[htbp]
\centering
\begin{tabular}{llll}
\toprule
fit range & food\_rest & energy & other \\
\midrule
2011Q2 - 2021Q2 & \phantom{-}0.32$^{***}$ & \phantom{-}0.05$^{**}$\phantom{$^{*}$} & \phantom{-}0.11\phantom{$^{***}$} \\
2006Q2 - 2021Q2 & \phantom{-}0.30$^{***}$ & -0.01\phantom{$^{***}$} & \phantom{-}0.51$^{***}$ \\
2002Q1 - 2021Q2 (max) & \phantom{-}0.31$^{***}$ & -0.02$^{*}$\phantom{$^{**}$} & \phantom{-}0.67$^{***}$ \\
%2016Q2 - 2021Q2 & \phantom{-}0.06\phantom{$^{***}$} & \phantom{-}0.02\phantom{$^{***}$} & \phantom{-}0.20\phantom{$^{***}$} \\
\bottomrule
\end{tabular}
\caption{Regression coefficients for responsiveness of mean IAS inflation perceptions to CPIH sub-component inflation for different time periods. {*}{*}{*} indicates significance at the 1\% level, {*}{*} at the 5\%
level, and {*} at the 10\% level. Robust standard errors have been used.}
\label{tab:compo_coeffs}
\end{table}

\section{Discussion}
\label{sec:discussion}

% summary
We investigated an LLM's ability to form perceptions of current and expectations of future consumer price inflation based on a set of economic conditioning information in a representative survey setting. Our analysis provided a set of tools to assess model sensitivity and explainability with respect to the conditioning information. We found that the used GPT model can reproduce key characteristics of human responses and official statistics. This is an impressive feat, as GPT models have not been explicitly designed to perform the tasks investigated here and no specific fine tuning was applied to the model: we achieved partially satisfactory performance out-of-the box for a relatively complex problem.\\
% shortcomings
However, our analysis also revealed several shortcomings of the used model in the given context. Similarly to human responses, GPT overweights the importance of salient price components, like those for food. This could be an advantage depending on the application. However, it also partially shows no sensitivity to the main other component. Its sensitivity exhibits strong non-linearities even within historically observed ranges of a kink-type nature for no apparent reason. As such, GPT does not appear to have a fully consistent world model for consumer price inflation. Additionally, GPT responses' correlation with those of actual survey respondents is relatively weak. This latter observation may largely be caused by the very limited conditioning information we provided to the model, which will certainly be insufficient to capture the idiosyncratic state of human respondents well. Achieving better alignment on the micro level is an interesting research problem in itself leaving plenty of scope for future work.\\
% discussion
We conclude our analysis with a discussion of ethical consideration when using LLMs to simulate or investigate human behaviour. One concern may be the simulation of individual persons, here survey subjects, without their explicit consent. This aspect needs to be addressed by data governance and is independent of the model type being used. The same concern applies to a linear regression model as much as it does to a more complex neural network, such as an LLM.\\
However, the more complex nature of LLMs leads us to a second concern which is more difficult to address. Namely, the black box critique of machine learning models, which is even more aggravated in the LLM case compared to supervised learning approaches. Model complexity (like the number of parameters in the neural network) is orders of magnitude larger compared to those used in supervised problems. While the training data (often vast corpora of unstructured data) are mostly unknown and inaccessible to the user, as are the details of the training algorithms used to build the model.\\
This second point puts more burden on model validation and testing. Much of this paper was dedicated to just that. However, as we have seen the comparison with human benchmark data (actual survey responses) and official statistics can still lead to difficult questions, because these three may not be aligned. LLM outputs may match human benchmark data, but fail to reproduce other statistics, or vice versa. There are many reasons for why this can happen, and it may not be possible to address these in the analysis leaving us with the difficult decision of whether and how to use LLM results, for instance to inform decisions.\\
Considering the goal of the analysis can serve as guidance here. If the aim is to replicate or analyse official statistics, such as macroeconomic aggregates, mismatch on more disaggregate levels can be tolerated if validation has been sufficiently satisfactory by some criteria. As pointed out, mismatch on the micro level does not preclude match on the aggregate level, and it may actually be a source of better aggregate performance.\\
However, if the goal of the analysis is to infer characteristics of humans or subgroups of the population, like demographic groups, potentially informing decisions affecting them, mismatch may not be tolerable if it cannot be accounted for in some way, such as via bias correction when using LLM outputs in a downstream model \citep{ludwig2025llm}. A decision on the usefulness of LLM outputs will most likely face trade-offs even after validation and possible corrections, necessitating a robust governance structure for the use of AI in decision making.\\
Finally, most of our results where obtained for a single model without a guarantee that these generalise to other models or settings. Given the complexity and diversity of LLMs, it is therefore essential to validate a model whenever either the model or setting changes as long as we do not have a general understanding of LLM behaviour in a given context. However, if validation has been satisfactory, the use of LLMs (or a particular LLM) opens new doors for research given their ability to simulate complex entities, like survey subjects, in-situ, cheaply and flexibly in a way which would be impossible in real-world settings. 

\newpage

\section*{Appendix}

\subsection*{Additional results}

% >>> time profile plot
\begin{figure}[h]
\centering
\includegraphics[width=.7\textwidth]{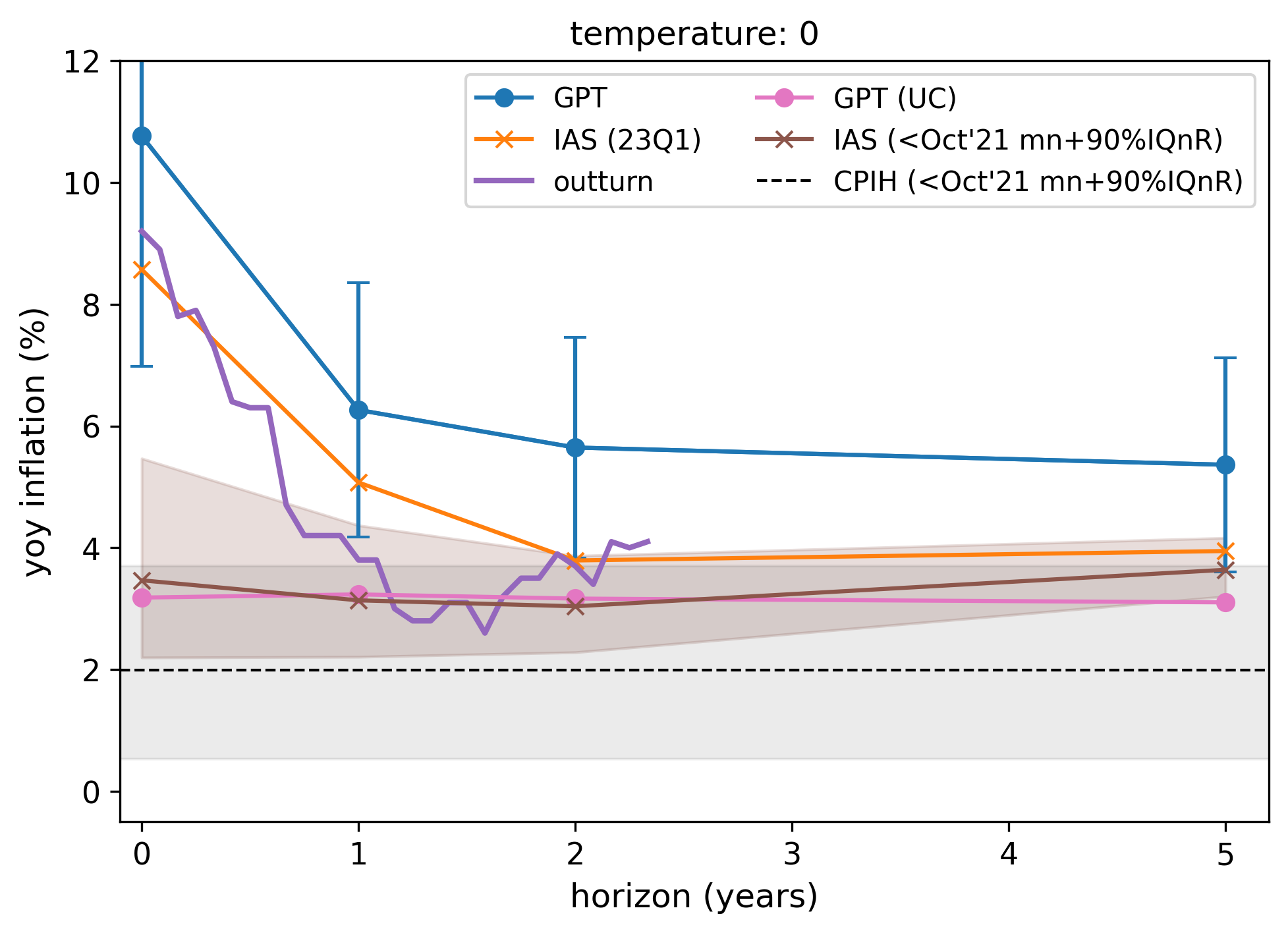}
\caption{Time profile of aggregate inflation expectations for IAS and GPT responses ($T=0$) for the main sample (2023Q1 survey): GPT conditioned (blue), GPT unconditioned (pink), IAS (sand), IAS pre-Oct-2021 mean \& swath (brown), out-turn from February 2023 onward (purple), and CPIH pre-Oct-2021 mean \& swath. Sources: IAS, ONS, authors' calculations.}
\label{fig:time_profile_T0}
\end{figure}

% >>> regression plots (housing, income, age)

\newpage

\subsection*{GPT knowledge validation}

Validation prompts and answers of \texttt{gpt-3.5-turbo-0613} as of 1.\ December\ 2025. Each question was presented in a new chat.\\[.2cm]

\texttt{Prompt:} Do you know the Bank of England's Inflation Attitudes Survey?    If yes, please briefly describe it and the latest results you remember.\\
\texttt{GPT:} Yes, I am familiar with the Bank of England's Inflation Attitudes Survey. It is a quarterly survey conducted by the Bank of England to gauge public attitudes and expectations towards inflation in the United Kingdom. The survey collects data on various aspects such as people's perceptions of current and future inflation rates, their understanding of inflation, and their expectations for future interest rates. As an AI language model, I don't have real-time data access, so I can't provide the latest results. However, the Bank of England publishes the survey results on its website, and you can find the most recent data there.\\[.2cm]

\texttt{Prompt:} What is the latest date you have data from Bank of England's Inflation Attitudes Survey?\\
\texttt{GPT:} As an AI language model, I don't have real-time data. The latest available data from the Bank of England's Inflation Attitudes Survey is from the May 2021 release. However, it's always best to refer to the Bank of England's official website or publications for the most up-to-date information.\\[.2cm]

\texttt{Prompt:} When did the Russian special military operation in Ukraine start?\\
\texttt{GPT:} The Russian special military operation in Ukraine began in February 2014. It started with the annexation of Crimea by Russia, followed by the support of separatist movements in eastern Ukraine.\\[.2cm]

\texttt{Prompt:} What is the latest event you can recall and its date?\\
\texttt{GPT:} As an AI language model, I don't have real-time information or the ability to recall specific events. My responses are based on pre-existing knowledge up until September 2021.\\[.2cm]

\texttt{Prompt:} Do you know what UK CPIH inflation was in February 2023? If not, say so, and make a guess.\\
\texttt{GPT:} I'm sorry, but as an AI language model, I don't have access to real-time data or the ability to browse the internet. Therefore, I cannot provide you with the specific UK CPIH inflation rate for February 2023. It would be best to consult official sources such as the Office for National Statistics (ONS) or financial news outlets for the most up-to-date information.\\[.2cm]

\texttt{Prompt:} Do you know what UK food price inflation was in February 2023? If not, say so, and make a guess.\\
\texttt{GPT:}I'm sorry, but as an AI language model, I don't have access to real-time data or the ability to browse the internet. Therefore, I cannot provide you with the specific food price inflation for February 2023 in the UK. It would be best to consult reliable sources or official statistics for the most accurate and up-to-date information on food price inflation.\\[.2cm]

\texttt{Prompt:} Do you know what UK energy price inflation was in February 2023? If not, say so, and make a guess.\\
\texttt{GPT:} I'm sorry, but as an AI language model, I don't have access to real-time data or the ability to browse the internet. Therefore, I cannot provide you with the specific energy price inflation for February 2023 in the UK. It would be best to consult reliable sources or financial institutions for the most up-to-date information on energy price inflation.\\[.2cm]

\texttt{Prompt:} Do you know what UK restaurant price inflation was in February 2023? If not, say so, and make a guess.\\
\texttt{GPT:} I'm sorry, but as an AI language model, I don't have access to real-time data or the ability to browse the internet. Therefore, I cannot provide you with the specific UK restaurant price inflation for February 2023. It would be best to consult official sources or financial reports for the most accurate and up-to-date information.\\[.2cm]

\texttt{Prompt:} What is the latest UK CPIH inflation number and date you recall?\\
\texttt{GPT:} As an AI language model, I don't have real-time data. The latest UK Consumer Prices Index including owner occupiers' housing costs (CPIH) inflation number I recall is 2.0\% for the year ending July 2021 [WRONG]. Please note that this information may not be up to date, so it's always best to refer to official sources like the Office for National Statistics (ONS) for the most recent data.\\[.2cm]

\texttt{Prompt:} Do you recall what UK CPIH inflation was in February 2021?\\
Yes, I can provide you with that information. The UK CPIH inflation rate in February 2021 was 0.7\% [CORRECT].\\[.2cm]

\texttt{Prompt:} Do you recall what UK food price inflation was in February 2021? If not, make a guess.\\
\texttt{GPT:} I'm sorry, but as an AI language model, I don't have real-time data or the ability to recall specific information like recent food price inflation in the UK. My responses are based on general knowledge up until September 2021. It's best to consult reliable sources or conduct a search to find the most up-to-date information on UK food price inflation in February 2021.

\texttt{Prompt:} Do you recall what UK energy price inflation was in February 2021? If not, make a guess.\\
\texttt{GPT:} I'm sorry, but as an AI language model, I don't have real-time data or the ability to recall specific information like energy price inflation in February 2021. My responses are based on general knowledge up until September 2021. It's always best to refer to reliable sources or consult the latest data from official sources for accurate and up-to-date information.\\[.2cm]

\texttt{Prompt:} Do you recall what UK restaurant price inflation was in February 2021? If not, make a guess.\\
\texttt{GPT:} I'm sorry, but as an AI language model, I don't have access to real-time data or the ability to recall specific information from a given time period. Therefore, I cannot provide you with the exact UK restaurant price inflation in February 2021. It would be best to consult official sources or financial reports for accurate and up-to-date information.

\subsection*{Demographic groups}\label{app_demo_groups}

\begin{table}[htbp]
\centering
\tiny
\begin{tabular}{lll}
\toprule
IAS number & Wording for prompt [IAS wording, if different] & ONS name(s)  \\
\midrule
Age \\
\midrule
2 & 25-34 &  Less than 30 (1/3), 30 to 49 (1/4)\\
3 & 35-44 & 30 to 49 (1/2)\\
4 & 45-54 & 30 to 49 (1/4), 50 to 64 (1/3)\\
5 & 55-64 & 50 to 64 (2/3)\\
6 & 16-24 & Less than 30 (2/3)\\
7 & 65-75 & 65 to 74 \\
\midrule
Education \\
\midrule
1 & got your GCSEs but not A-levels [low (GCSE)] & Aged 16\\
2 & got your A-levels but not a degree [medium (A-level)] & Aged 17 and under 19\\
3 & have got a degree [high (degree)] & Aged 19 and under 22\\
NaN & haven't shared your level of education & Aged 22 or over\\
\midrule
Housing tenure \\
\midrule
1 & house you own outright [Owned outright] & Owners Owned outright\\
2 & house with a mortgage [Mortgage] & Buying with a mortgage\\
3 & council house [Council] & Social rented from Council\\
4 & house you rent [Other] & Registered Social Landlord \\
&&( + all other rental fields)\\
\midrule
Income \\
\midrule
7 & income of <£9999 & Lowest ten (4/5)\\
8 & income of £10000-£19999 & Lowest ten (1/5), Second decile, Third decile (1/2)\\
9 & income of £20000-£34999 & Third decile (1/2), Fourth decile, Fifth decile\\
10 & income of £35000-£44999 & Sixth decile, Seventh decile (1/2)\\
11 & income of >£45000 & Seventh decile (1/2), Eighth decile, Ninth decile, \\
&&Highest ten\\
12 & undisclosed income [Prefer not to answer] \\
NaN & undisclosed income\\
\midrule
Region \\
\midrule
1 & Scotland & Scotland\\ 
2 & the North of England or Northern Ireland & North East, North West, \\
&&Yorkshire and The Humber, Northern Ireland\\
3 & the Midlands (of England) [the Midlands] & East Midlands, West Midlands, East\\
4 & Wales or the West of England & South West, Wales\\
5 & the South East of England [the South East] & London, South East\\
\midrule
Sex \\
\midrule
1 & male \\
2 & female \\
3 & identifying your gender in another way [In another way]\\
4 & not revealing your gender [Prefer not to answer] \\
\midrule
Social class \\
\midrule
1 & upper-middle class & Large employers and higher managerial, \\
&&Higher professional\\
2 & lower middle class & Lower managerial and professional, \\
&& Intermediate, Small employers\\
3 & skilled working class & Lower supervisory, Semi-routine\\
4 & working class & Routine\\
& are a pensioner \\
\midrule
Work \\
\midrule
1 & work full or part time [Full or Part time]\\
2 & are not working [Not working] \\
\bottomrule
\end{tabular}
\caption{The demographic groups, showing their numbering in the IAS survey, rendering in the prompt and the IAS survey (if different) and the mapping of ONS groups to the IAS demographics.}\label{tab:demo_list}
\end{table}

Table \ref{tab:demo_list} lists the demographic groups included in the analysis. The different demographic groups are represented by a number code in the IAS (left-hand column). We map those codes to a human-readable string in the prompt: these are closely based on the IAS names however are clarified if required to stand alone, or to make it clear that we are referring to the UK (for example, `the South East', does not distinguish from the South East US). We also report the mapping from the ONS demographic groups, which often do not line up with the IAS definitions. When these groups are distributed across IAS groups, we show the fractions we use in brackets. We impute `are a pensioner' as a social class based on age, since otherwise those individuals show up as an unemployed member of the social class of their previous job.\\[.2cm]

\bibliographystyle{apalike}
\bibliography{references}

@article{brown2020language,
  title={Language models are few-shot learners},
  author={Brown, Tom and Mann, Benjamin and Ryder, Nick and Subbiah, Melanie and Kaplan, Jared D and Dhariwal, Prafulla and Neelakantan, Arvind and Shyam, Pranav and Sastry, Girish and Askell, Amanda and others},
  journal={Advances in neural information processing systems},
  volume={33},
  pages={1877--1901},
  year={2020}
}

@article{ouyang2022training,
  title={Training language models to follow instructions with human feedback},
  author={Ouyang, Long and Wu, Jeffrey and Jiang, Xu and Almeida, Diogo and Wainwright, Carroll and Mishkin, Pamela and Zhang, Chong and Agarwal, Sandhini and Slama, Katarina and Ray, Alex and others},
  journal={Advances in neural information processing systems},
  volume={35},
  pages={27730--27744},
  year={2022}
}

@inproceedings{pezeshkpour2024large,
  title={Large language models sensitivity to the order of options in multiple-choice questions},
  author={Pezeshkpour, Pouya and Hruschka, Estevam},
  booktitle={Findings of the Association for Computational Linguistics: NAACL 2024},
  pages={2006--2017},
  year={2024}
}

@article{Lucas1975,
  title={An equilibrium model of the business cycle},
  author={Lucas, Robert},
  journal={{Journal of Political Economy}},
  volume={83},
  number={6},
  pages={1113--1144},
  year={1975},
  publisher={University of Chicago Press}
}

@article{DAcuntoWeber2024,
  title={Why Survey-Based Subjective Expectations Are Meaningful and Important},
  author={D'Acunto, Francesco and Weber, Michael},
  journal={Annual Review of Economics},
  volume={16},
  pages={329-357},
  year={2024}
}

@article{Bernanke2007,
  title={Inflation expectations and inflation forecasting},
  author={Bernanke, Ben},
  journal={Speech at the Monetary Economics Workshop of the National Bureau of Economic Research Summer Institute, Cambridge, Massachusetts},
  year={2007}
}

@article{MankiwReis2002,
  title={Sticky information versus sticky prices: a proposal to replace the New Keynesian Phillips curve},
  author={Mankiw, Gregory and Reis, Ricardo},
  journal={The Quarterly Journal of Economics},
  volume={117},
  number={4},
  pages={1295--1328},
  year={2002},
  publisher={Oxford University Press}
}

@article{MackowiakWiederholt2009,
  title={Optimal sticky prices under rational inattention},
  author={Mackowiak, Bartosz and Wiederholt, Mirko},
  journal={American Economic Review},
  volume={99},
  number={3},
  pages={769--803},
  year={2009}
}

@article{CoibionGorodnichenko2015a,
  title={Information rigidity and the expectations formation process: A simple framework and new facts},
  author={Coibion, Olivier and Gorodnichenko, Yuriy},
  journal={American Economic Review},
  volume={105},
  number={8},
  pages={2644--78},
  year={2015}
}

@article{VanderKlaauw2012,
  title={The formation of inflation expectations},
  author={Van der Klaauw, Wilbert and Bruine de Bruin, W{\"a}ndi and Topa, Giorgio and Potter, Simon and Bryan, Michael},
  journal={Annual Review of Economics},
  volume={4},
  number={1},
  pages={381--405},
  year={2012},
  publisher={Annual Reviews}
}

@article{deBruin2011,
  title={Expectations of inflation: The role of demographic variables, expectation formation, and financial literacy},
  author={De Bruin, W{\"a}ndi Bruine and Van der Klaauw, Wilbert and Downs, Julie and Fischhoff, Baruch and Topa, Giorgio and Armantier, Olivier},
  journal={Journal of Consumer Affairs},
  volume={45},
  number={3},
  pages={381--402},
  year={2011},
  publisher={Wiley Online Library}
}

@article{DAcunto2021,
  title={Exposure to grocery prices and inflation expectations},
  author={D'Acunto, Francesco and Malmendier, Ulrike and Ospina, Juan and Weber, Michael},
  journal={Journal of Political Economy},
  volume={129},
  number={5},
  pages={1615--1639},
  year={2021},
  publisher={University of Chicago Press Chicago, IL}
}

@article{Mankiw2003,
author = {Mankiw, N. Gregory and Reis, Ricardo and Wolfers, Justin},
title = {Disagreement about Inflation Expectations},
journal = {NBER Macroeconomics Annual},
volume = {18},
number = {},
pages = {209-248},
year = {2003}
}

@article{Aher2022,
  title={Using large language models to simulate multiple humans},
  author={Aher, Gati and Arriaga, Rosa and Kalai, Adam Tauman},
  journal={arXiv},
  volume={2208.10264},
  year={2022}
}

@article{Brookins2023,
  title={Playing Games With {GPT}: What Can We Learn About a Large Language Model From Canonical Strategic Games?},
  author={Brookins, Philip and DeBacker, Jason Matthew},
  journal={Available at SSRN 4493398},
  year={2023}
}

@article{Bybee2023,
  title={Surveying Generative AI's Economic Expectations},
  author={Bybee, Leland},
  journal={arXiv},
  volume={2305.02823},
  year={2023}
}

@techreport{FariaeCastro2023,
  title={Artificial Intelligence and Inflation Forecasts},
  author={Faria-e-Castro, Miguel and Leibovici, Fernando},
  number={2023-015},
  year={2023},
  institution={Federal Reserve Bank of St. Louis}
}

@techreport{Horton2023,
  title={{Large language models as simulated economic agents: What can we learn from homo silicus?}},
  author={Horton, John},
  number={31122},
  year={2023},
  institution={National Bureau of Economic Research}
}

@article{Jiang2022,
  title={Chain-of-thought prompting elicits reasoning in large language models},
  author={Jiang, Wenlong and others},
  journal={Advances in Neural Information Processing Systems},
  volume={35},
  pages={24824--24837},
  year={2022}
}

@article{StrumbeljKononenko2010,
  title={An efficient explanation of individual classifications using game theory},
  author={Strumbelj, Erik and Kononenko, Igor},
  journal={Journal of Machine Learning Research},
  volume={11},
  number={Jan},
  pages={1--18},
  year={2010}
}

@article{Joseph2019,
  title={From interpretability to inference: an estimation framework for universal approximators}, 
      author={Andreas Joseph},
      year={2019},
      volume={1903.04209},
      journal={arXiv}
}

@article{BuckmanJoseph2022,
  journal={International Journal of Central Banking},
author={Marcus Buckmann and Andreas Joseph},
title={An Interpretable Machine Learning Workflow with an Application to Economic Forecasting},
year={2023},
month={October},
pages={449-522},
volume={19},
number={4}
}

@article{MalmendierNagel2013,
  title={Learning from inflation experiences},
  author={Malmendier, Ulrike and Nagel, Stefan},
  journal={The Quarterly Journal of Economics},
  volume={131},
  number={1},
  pages={53--87},
  year={2016},
  publisher={Oxford University Press}
}

@article{Krause2011SI,
title = {Swarm intelligence in humans: diversity can trump ability},
journal = {Animal Behaviour},
volume = {81},
number = {5},
pages = {941-948},
year = {2011},
issn = {0003-3472},
doi = {https://doi.org/10.1016/j.anbehav.2010.12.018},
url = {https://www.sciencedirect.com/science/article/pii/S0003347210005221},
author = {Krause, Stefan and James, Richard and Faria, Jolyon and Ruxton, Graeme and Krause, Jens},
}

@article{Galton1907VP,
  author = {F. Galton},
  journal = {Nature},
  number = {7},
  pages = {450--1},
  title = {Vox Populi (the Wisdom of Crowds)},
  volume = {75},
  year = {1907}
}

@article{ludwig2025llm,
      title={Large Language Models: An Applied Econometric Framework}, 
      author={Jens Ludwig and Sendhil Mullainathan and Ashesh Rambachan},
      year={2025},
      volume={2412.07031},
      journal={arXiv}
}

@article{christie2024judgement,
  title={{Large language models can outperform humans in social situational judgments}},
  url={https://doi.org/10.1038/s41598-024-79048-0},
  doi={10.1038/s41598-024-79048-0},
  journal={{Scientific Reports}},
  author={Christie, Fiona and O'Connor, Sarah and Kruschwitz, Walter},
  year={2024},
  volume={14},
  number={27449}
}

@techreport{galatzer2024benchmark,
      title={{The Cognitive Capabilities of Generative AI: A Comparative Analysis with Human Benchmarks}}, 
      author={Isaac R. Galatzer-Levy and David Munday and Jed McGiffin and Xin Liu and Danny Karmon and Ilia Labzovsky and Rivka Moroshko and Amir Zait and Daniel McDuff},
      year={2024},
      eprint={2410.07391},
      institution={arXiv},
      primaryClass={cs.AI},
      url={https://arxiv.org/abs/2410.07391}, 
}

@article{luo2024neuro,
   title={Large language models surpass human experts in predicting neuroscience results},
   volume={9},
   ISSN={2397-3374},
   url={http://dx.doi.org/10.1038/s41562-024-02046-9},
   DOI={10.1038/s41562-024-02046-9},
   number={2},
   journal={Nature Human Behaviour},
   publisher={Springer Science and Business Media LLC},
   author={Luo, Xiaoliang and Rechardt, Akilles and Sun, Guangzhi and Nejad, Kevin K. and Yáñez, Felipe and Yilmaz, Bati and Lee, Kangjoo and Cohen, Alexandra O. and Borghesani, Valentina and Pashkov, Anton and Marinazzo, Daniele and Nicholas, Jonathan and Salatiello, Alessandro and Sucholutsky, Ilia and Minervini, Pasquale and Razavi, Sepehr and Rocca, Roberta and Yusifov, Elkhan and Okalova, Tereza and Gu, Nianlong and Ferianc, Martin and Khona, Mikail and Patil, Kaustubh R. and Lee, Pui-Shee and Mata, Rui and Myers, Nicholas E. and Bizley, Jennifer K. and Musslick, Sebastian and Bilgin, Isil Poyraz and Niso, Guiomar and Ales, Justin M. and Gaebler, Michael and Ratan Murty, N. Apurva and Loued-Khenissi, Leyla and Behler, Anna and Hall, Chloe M. and Dafflon, Jessica and Bao, Sherry Dongqi and Love, Bradley C.},
   year={2024},
   month=nov, pages={305–315} }

@article{licorish2025comparing,
  title={Comparing Human and {LLM} Generated Code: The Jury is Still Out!},
  author={Licorish, Sherlock A and Bajpai, Ansh and Arora, Chetan and Wang, Fanyu and Tantithamthavorn, Chakkrit},
  journal={arXiv},
  volume={2501.16857},
  year={2025}
}

@article{bai2025bias,
author = {Xuechunzi Bai  and Angelina Wang  and Ilia Sucholutsky  and Thomas L. Griffiths },
title = {Explicitly unbiased large language models still form biased associations},
journal = {Proceedings of the National Academy of Sciences},
volume = {122},
number = {8},
pages = {e2416228122},
year = {2025},
doi = {10.1073/pnas.2416228122},
URL = {https://www.pnas.org/doi/abs/10.1073/pnas.2416228122},
eprint = {https://www.pnas.org/doi/pdf/10.1073/pnas.2416228122}
}

@article{rubin2005outcomes,
 author = {Donald B. Rubin},
 journal = {Journal of the American Statistical Association},
 number = {469},
 pages = {322--331},
 publisher = {[American Statistical Association, Taylor & Francis, Ltd.]},
 title = {Causal Inference Using Potential Outcomes: Design, Modeling, Decisions},
 urldate = {2025-11-28},
 volume = {100},
 year = {2005}
}

@TechReport{repec:boe:boeewp:1125,
type={Bank of England, Staff Working Paper},
institution={Bank of England},
author={Nikoleta Anesti and Vania Esady and Matthew Naylor},
title={Food prices matter most: sensitive household inflation expectations},
year={2025},
month={Apr},
number={1125},
doi={None},
url={https://ideas.repec.org/p/boe/boeewp/1125.html},
}

@techreport{hobijn2009household,
  type={Working Paper Series},
institution={Federal Reserve Bank of San Francisco},
author={Bart Hobijn and Kristin Mayer and Carter Stennis and Giorgio Topa},
title={{Household inflation experiences in the U.S.: a comprehensive approach}},
year={2009},
number={2009-19}
}

@article{kaplan2017inflation,
  title={{Inflation at the Household Level}},
  author={Kaplan, Greg and Schulhofer-Wohl, Sam},
  journal={Journal of Monetary Economics},
  volume={91},
  pages={19--38},
  year={2017},
  publisher={Elsevier}
}

@techreport{arioli2017eu,
  title={{EU Consumers' Quantitative Inflation Perceptions and Expectations: an Evaluation}},
  author={Arioli, Rodolfo and Bates, Colm and Dieden, Heinz and Duca, Ioana and Friz, Roberta and Gayer, Christian and Kenny, Geoff and Meyler, Aidan and Pavlova, Iskra},
  year={2017},
number = {186},
  institution={ECB Occasional Paper}
}

@book{del2008s,
  title={{What's behind “Inflation Perceptions”? A Survey-based Analysis of Italian Consumers}},
  author={Del Giovane, Paolo and Fabiani, Silvia and Sabbatini, Roberto},
  year={2008},
  publisher={Springer}
}

@article{jonung1981perceived,
  title={{Perceived and Expected Rates of Inflation in Sweden}},
  author={Jonung, Lars},
  journal={The American Economic Review},
  volume={71},
  number={5},
  pages={961--968},
  year={1981},
  publisher={JSTOR}
}

@article{de2011measuring,
  title={{Measuring Consumer Uncertainty about Future Inflation}},
  author={Bruine de Bruin, W{\"a}ndi and Manski, Charles F and Topa, Giorgio and Van Der Klaauw, Wilbert},
  journal={Journal of Applied Econometrics},
  volume={26},
  number={3},
  pages={454--478},
  year={2011},
  publisher={Wiley Online Library}
}

@article{cavallo2017inflation,
  title={{Inflation Expectations, Learning, and Supermarket Prices: Evidence from Survey Experiments}},
  author={Cavallo, Alberto and Cruces, Guillermo and Perez-Truglia, Ricardo},
  journal={American Economic Journal: Macroeconomics},
  volume={9},
  number={3},
  pages={1--35},
  year={2017},
  publisher={American Economic Association 2014 Broadway, Suite 305, Nashville, TN 37203-2425}
}

@article{COIBION2020103297,
title = {{Inflation Expectations as a Policy Tool?}},
journal = {Journal of International Economics},
volume = {124},
pages = {103297},
year = {2020},
doi = {https://doi.org/10.1016/j.jinteco.2020.103297},
url = {https://www.sciencedirect.com/science/article/pii/S0022199620300167},
author = {Olivier Coibion and Yuriy Gorodnichenko and Saten Kumar and Mathieu Pedemonte},
}

@article{coibion2022monetary,
  title={{Monetary Policy Communications and Their Effects on Household Inflation Expectations}},
  author={Coibion, Olivier and Gorodnichenko, Yuriy and Weber, Michael},
  journal={Journal of Political Economy},
  volume={130},
  number={6},
  pages={1537--1584},
  year={2022},
  publisher={The University of Chicago Press Chicago, IL}
}

@techreport{d2020effective,
  title={{Effective Policy Communication: Targets versus Instruments}},
  author={D'Acunto, Francesco and Hoang, Daniel and Paloviita, Maritta and Weber, Michael},
  institution={Chicago Booth Research Paper},
  number={20-38},
  year={2020}
}

@inproceedings{d2019cognitive,
  title={{Cognitive Abilities and Inflation Expectations}},
  author={D'Acunto, Francesco and Hoang, Daniel and Paloviita, Maritta and Weber, Michael},
  booktitle={AEA Papers and Proceedings},
  volume={109},
  pages={562--566},
  year={2019}
}

@article{ehrmann2022central,
  title={{Central Bank Communication with Non-Experts--A Road to Nowhere?}},
  author={Ehrmann, Michael and Wabitsch, Alena},
  journal={Journal of Monetary Economics},
  volume={127},
  pages={69--85},
  year={2022},
  publisher={Elsevier}
}

@article{lamla2015information,
  title={{Information Rigidities, Inflation Perceptions, and the Media: Lessons from the Euro Cash Changeover}},
  author={Lamla, Michael J and Lein, Sarah M},
  journal={Economic Inquiry},
  volume={53},
  number={1},
  pages={9--22},
  year={2015},
  publisher={Wiley Online Library}
}

@article{macaulay2022heterogeneous,
  title={{Heterogeneous Information, Subjective Model Beliefs, and the Time-Varying Transmission of Shocks}},
  author={Macaulay, Alistair},
  year={2022},
  Journal={CESifo Working Paper}
}

@article{mcmahon2023getting,
  title={{Getting Through: Communicating Complex Information}},
  author={McMahon, Michael and Naylor, Matthew},
  year={2023},
volume={1047},
  journal={Bank of England Staff Working Paper}
}

@article{sims2010rational,
  title={{Rational Inattention and Monetary Economics}},
  author={Sims, Christopher A},
  journal={Handbook of Monetary Economics},
  volume={3},
  number={C},
  pages={155--181},
  year={2010},
  publisher={Elsevier}
}

@article{Zarifhonarvar2025survey,
title = {Generating inflation expectations with large language models},
journal = {Journal of Monetary Economics},
volume = {157},
pages = {103859},
year = {2026},
doi = {https://doi.org/10.1016/j.jmoneco.2025.103859},
author = {Ali Zarifhonarvar}
}

@article{Hansen2025Simulating,
  title={Simulating the Survey of Professional Forecasters},
  author={Hansen, Anne Lundgaard and Horton, John J and Kazinnik, Sophia and Puzzello, Daniela and Zarifhonarvar, Ali},
  journal={SSRN 5066286},
  year={2025},
  publisher={SSRN}
}

@article{Acemoglu2022labour,
author = {Acemoglu, Daron and Autor, David and Hazell, Jonathon and Restrepo, Pascual},
title = {Artificial Intelligence and Jobs: Evidence from Online Vacancies},
journal = {Journal of Labor Economics},
volume = {40},
number = {S1},
pages = {S293-S340},
year = {2022}
}

@article{korinek2023ai,
  author  = {Korinek, Anton},
  title   = {{Generative AI for Economic Research: Use Cases and Implications for Economists}},
  journal = {Journal of Economic Literature},
  year    = {2023},
  volume  = {61},
  number  = {4},
  pages   = {1281-1317},
  doi     = {10.1257/jel.20231736}
}

@article{korinek2025ai,
  author = {Korinek, Anton},
  title = {{AI Agents for Economic Research}},
  journal = {Journal of Economic Literature},
  year = {2025},
  url = {https://www.aeaweb.org/articles/materials/23826}
}

@techreport{chen2024displacement,
  author = {Chen, Wilbur Xinyuan and Srinivasan, Suraj and Zakerinia, Saleh},
  year = {2024},
  title = {{Displacement or Complementarity? The Labor Market Impact of Generative AI}},
  institution = {{Harvard Business School Working Paper, No. 25-039}}
}

@article{hui2024labour,
author = {Hui, Xiang and Reshef, Oren and Zhou, Luofeng},
title = {The Short-Term Effects of Generative Artificial Intelligence on Employment: Evidence from an Online Labor Market},
year = {2024},
volume = {35},
number = {6},
issn = {1526-5455},
url = {https://doi.org/10.1287/orsc.2023.18441},
doi = {10.1287/orsc.2023.18441},
journal = {Organization Science},
pages = {1977–1989}
}

@inproceedings{arora2023probing,
    title = "Probing Pre-Trained Language Models for Cross-Cultural Differences in Values",
    author = "Arora, Arnav  and
      Kaffee, Lucie-aim{\'e}e  and
      Augenstein, Isabelle",
    editor = "Dev, Sunipa  and
      Prabhakaran, Vinodkumar  and
      Adelani, David Ifeoluwa  and
      Hovy, Dirk  and
      Benotti, Luciana",
    booktitle = "Proceedings of the First Workshop on Cross-Cultural Considerations in NLP (C3NLP)",
    month = may,
    year = "2023",
    address = "Dubrovnik, Croatia",
    publisher = "Association for Computational Linguistics",
    url = "https://aclanthology.org/2023.c3nlp-1.12/",
    doi = "10.18653/v1/2023.c3nlp-1.12",
    pages = "114--130",
   }

@article{Argyle2023simulate, 
title={Out of One, Many: Using Language Models to Simulate Human Samples}, 
volume={31}, 
DOI={10.1017/pan.2023.2}, 
number={3}, 
journal={Political Analysis}, 
author={Argyle, Lisa P. and Busby, Ethan C. and Fulda, Nancy and Gubler, Joshua R. and Rytting, Christopher and Wingate, David}, 
year={2023}, 
pages={337–351}
}

@article{ross2024llmeconomicusmappingbehavioral,
      title={{LLM economicus? Mapping the Behavioral Biases of LLMs via Utility Theory}}, 
      author={Jillian Ross and Yoon Kim and Andrew W. Lo},
      year={2024},
      volume={2408.02784},
      journal={arXiv},
      primaryClass={cs.CL},
      url={https://arxiv.org/abs/2408.02784}, 
}

@article{henning2025llmagentsreplicatehuman,
      title={{LLM Agents Do Not Replicate Human Market Traders: Evidence From Experimental Finance}}, 
      author={Thomas Henning and Siddhartha M. Ojha and Ross Spoon and Jiatong Han and Colin F. Camerer},
      year={2025},
      volume={2502.15800},
      journal={arXiv}
}

@article{charness2025next,
  title={The next generation of experimental research with LLMs},
  author={Charness, Gary and Jabarian, Brian and List, John A},
  journal={Nature Human Behaviour},
  volume={9},
  number={5},
  pages={833--835},
  year={2025},
  publisher={Nature Publishing Group UK London},
  doi={10.1038/s41562-025-02137-1}
}

@article{manning2024automatedsocialsciencelanguage,
      title={{Automated Social Science: Language Models as Scientist and Subjects}}, 
      author={Benjamin S. Manning and Kehang Zhu and John J. Horton},
      year={2024},
      volume={2404.11794},
      journal={arXiv}
}

@article{lopezlira2025memorizationproblemtrustllms,
      title={The Memorization Problem: Can We Trust LLMs' Economic Forecasts?}, 
      author={Alejandro Lopez-Lira and Yuehua Tang and Mingyin Zhu},
      year={2025},
      volume={2504.14765},
      journal={arXiv}
}

@article{orgad2025llmsknowshowintrinsic,
      title={{LLMs Know More Than They Show: On the Intrinsic Representation of LLM Hallucinations}}, 
      author={Orgad, Hadas and Toker, Michael and Gekhman, Zorik and Reichart, Roi and Szpektor, Idan and Kotek, Hadas and Belinkov, Yonatan Belinkov},
      year={2025},
      volume={2410.02707},
      journal={arXiv}
}

@techreport{buckmann2025,
  author = {Buckmann, Marcus and Nguyen, Quynh Anh and Hill, Ed},
  title = {Revealing economic facts: {LLMs} know more than they say},
  institution = {Bank of England},
  type = {Staff Working Paper},
  number = {1,150},
  year = {2025},
  month = {October},
  day = {31},
  url = {https://www.bankofengland.co.uk/-/media/boe/files/working-paper/2025/revealing-economic-facts-llms-know-more-than-they-say.pdf}
}

@techreport{buckmann2025improving,
  title={Improving Text Classification: Logistic Regression Makes Small LLMs Strong and Explainable'Tens-of-Shot'Classifiers},
  author={Buckmann, Marcus and Hill, Ed},
  institution = {Bank of England},
  type = {Staff Working Paper},
  number={1,127},
  year={2025},
  url = {https://www.bankofengland.co.uk/-/media/boe/files/working-paper/2025/improving-text-classification-logistic-regression-llms-tens-of-shot-classifiers.pdf}
}

@techreport{crane2025,
  author = {Crane, Leland Dod and Karra, Akhil and Soto, Paul E.},
  title = {Total Recall? {E}valuating the Macroeconomic Knowledge of Large Language Models},
  institution = {Board of Governors of the Federal Reserve System},
  series = {Finance and Economics Discussion Series (FEDS)},
  type = {Working Paper},
  number = {2025-44},
  year = {2025},
  month = {June},
  doi = {10.17016/FEDS.2025.044},
  url = {https://ssrn.com/abstract=5330692}
}

\end{document}